\pdfoutput=1
\documentclass[11pt]{article}

\usepackage{ACL2023}

\usepackage{times}
\usepackage{latexsym}
\usepackage{amsmath} % for \mathcal and advanced math
\usepackage{booktabs}
\usepackage{multirow} 
\usepackage{todonotes}
\usepackage{hyperref}
\usepackage{url}
\usepackage{fvextra}
\usepackage{CJKutf8}
\usepackage{multirow} 
\usepackage{tcolorbox}
\usepackage{siunitx}
\usepackage{amssymb}
\usepackage{enumitem}
\usepackage{microtype}
\usepackage{inconsolata}

\usepackage[T1]{fontenc}
\usepackage[utf8]{inputenc}

\usepackage{microtype}

\usepackage{inconsolata}

\newcommand{\round}[1]{\ensuremath{\lfloor#1\rceil}}
\title{FiRST: Finetuning Router-Selective Transformers for
Input-Adaptive Latency Reduction}
\newcommand{\AJ}[1]{\textcolor{black}{#1}}

\author{
    Akriti Jain\textsuperscript{1}$^*$, 
    Saransh Sharma\textsuperscript{2}$^*$,
    Koyel Mukherjee\textsuperscript{1}, 
    Soumyabrata Pal\textsuperscript{1} \\[0.5em]
    \textsuperscript{1}Adobe Research \\
    \textsuperscript{2}Indian Institute of Technology, Kharagpur \\[0.5em]
    \texttt{\{akritij, komukher, soumyabratap\}@adobe.com}, \texttt{saransh03sharma@kgpian.iitkgp.ac.in}
}

\begin{document}
\maketitle
\begin{abstract}
\def\thefootnote{*}\footnotetext{Equal Contribution}\def\thefootnote{\arabic{footnote}}
\footnotetext[2]{Work done during internship at Adobe Research}
Auto-regressive Large Language Models (LLMs) demonstrate remarkable performance across different domains such as vision and language tasks. However, due to sequential processing through multiple transformer layers, autoregressive decoding faces significant computational challenges, particularly in resource-constrained environments like mobile and edge devices. Existing approaches in literature that aim to improve latency via skipping layers have two distinct flavors: (1) early exit, and (2) input-agnostic heuristics where tokens exit at pre-determined layers irrespective of input sequence. Both the above strategies have limitations—the former cannot be applied in the presence of KV caching, which is essential for speed-ups in modern inference frameworks, and the latter fails to capture variation in layer importance across tasks or, more generally, across input sequences.   
To address these limitations, we propose \textsc{FiRST}, a model-agnostic framework that reduces inference latency by using layer-specific routers to adaptively skip transformer layers during decoding, based on routing decisions made from the input prompt in the prefill stage.  \textsc{FiRST} remains fully compatible with KV caching, enabling faster decoding while maintaining quality. 
Our method reveals that input adaptivity is essential—different tasks rely on different subsets of layers to evolve meaningful representations. Extensive experiments show that \textsc{FiRST} significantly reduces latency while outperforming existing layer selection strategies in quality. It retains performance comparable to the base model without skipping. 
\textsc{FiRST} is thus a promising and efficient solution for LLM deployment in low-resource environments.
\end{abstract}

\section{Introduction}

Large Language Models (LLMs) have revolutionized the fields of Natural Language Processing and Computer Vision achieving incredible performance on a diverse set of benchmark tasks. %For many downstream tasks, LLM's have provided a simplified and unified interface for inserting as input a small number of complex input/output pairs and achieving stellar performance. 
However, the massive scale of LLMs—often involving billions of parameters—poses significant challenges for deployment in resource-constrained environments, where memory, compute, and especially latency become critical bottlenecks. In this work, we focus on addressing the latency issue, which is particularly pronounced in edge settings such as laptops and mobile devices. As noted by \citet{schuster2022confident}, the auto-regressive nature of decoding in LLMs further amplifies this bottleneck.
%Our main interest lies in the resource-constrained on-device setting where resolving this bottleneck is of particular importance.

Transformer-based LLMs consist of multiple stacked layers—including attention and feed-forward networks—which result in high latency and computational cost. This makes inference slow or even impractical in resource-constrained environments. The inefficiency stems from the need to process each input token sequentially through all layers, irrespective of the input sequence or task. However, it is important to note that in the real world, there is a lot of heterogeneity in input sequences and tasks. \citep{schuster2022confident, sun2022simple} observed that LLM generations can have varying levels of difficulty and certain generations can be solved with reduced computation, by exiting the transformer stack early. At the same time, it has been noted in recent works \citep{llamas} that inference forward pass proceeds in phases through the layers of transformer-based models, with different types of information being extracted or mapped at different phases (sequences of layers) for certain tasks such as translation. Motivated by these and other related works, we hypothesize that \emph{different sequential combinations of layers are important for different input sequences and tasks}. Learning the right sequential combination of layers can help reduce inference latency and compute for on-device scenarios. However, there are several challenges. Any algorithm for determining the ``right'' combination of layers should minimize any quality loss, be compatible with other latency reduction strategies such as KV cache handling, and be learnable with minimal compute/training overhead. 
% i) \emph{Which combination of layers are important fora certain input and tasks?}, \newline 
% ii)\emph{Can such a determination be made at inference time?}, \newline
% iii) \emph{Can we learn with minimal compute and training overhead?}, \newline iv)\emph{How can we minimize quality loss?}\newline 

% to make LLM inferencing compatible wtih on-device constraints, we propo
% While transformer based LLM's have several stacks of layers (including attention and FFN layers), it has been observed \textcolor{red}{CITE} that the computational requirements for data-points vary significantly. %Furthermore, it has been observed that different layers of large models are important for different tasks. For example, for 
In the last few years, several promising approaches have been proposed in literature that adaptively prune layers at each decoding step. Token-level early exit proposed in \citep{schuster2022confident, sun2022simple} allow tokens to exit the transformer layer stack early based on different strategies to compute the confidence or saturation level. \citep{elhoushi2024layer, elbayad2019depth, zhang2019scan} extended this idea to incorporate layer skipping at a token level during training. % since pre-trained transformer layers spread their compute across layers. COULD WE EXPLAIN THIS? NOT VERY CLEAR.}
While token level early exit is a useful idea in theory, it suffers from a major limitation of incompatible KV caching in practice \citep{del2023skipdecode}. The incompatibility stems from having to recompute KV caches for preceding tokens if we have a delayed exit point for latter tokens, often resulting in loss of early exit advantages. This limits its practical adoption since KV cache is crucial in significantly speeding up auto-regressive decoding. %, %inappropriate handling of KV cache limits practical adoption. 

Recently, \citep{liu2024accelerating,del2023skipdecode,song2024sleb} have proposed  input-agnostic layer skipping at token level, that handle KV cache appropriately as well as retain the advantage of adaptive partial computation. In these solutions, tokens exit at pre-determined layers irrespective of the input sequence, and for all sequences in a batch, tokens at the same position in a sequence exit at the same layer. Furthermore, tokens at latter parts of the sequence are constrained to exit earlier than the previous tokens to ensure that there is no redundant KV cache re-computation. These solutions are heuristic based and impose hard rules and constraints irrespective of input sequences, which can lead to drop in output quality. Others \citep{jaiswal2024ffn, chen2024compressing} have proposed skipping layers by identifying redundant ones through computing cosine similarity of (input/output) representations of a layer. 
%circumventing the KV cache issue entirely by skipping only FFN layers.
%but such a strategy cannot reduce redundancy in transformer layer computations. Moreover, they propose an input adaptive skipping heuristic based on cosine similarity of outputs: if two adjacent layers have a similarity greater than a threshold,  then all subsequent layers except the last few are skipped. 
However, their strategy does not take into account that several middle layers are crucial (see \citep{liu2024accelerating}) and furthermore, final prediction capability of full model is not taken into account while deciding which layers to skip. \textit{Importantly, in none of the works described above, the strategy of selecting layers for skipping is sequence dependent. Furthermore, they do not consider fine-tuning the models in a way such that not only the performance improves but the model also learns to skip layers appropriately.}

\noindent Due to space constraints, we delegate a study of other related works and orthogonal approaches (for e.g. model compression) for exploring latency/performance tradeoff to Appendix \ref{app:related_work_appendix}.

Our goal is to design an input-adaptive, \textbf{learnable layer selection strategy} that provides quality-aware latency improvements while properly handling KV caching. Ideally, for each input sequence and task, we want to predict a sequential combination of layers to run during inference---minimizing quality loss while achieving as much latency gain as possible. We also want to do this with minimal computational overhead or extra training.
To achieve this, we propose training \textbf{routers}. At each layer, the router looks at the current sequence representation and decides whether to skip the next layer. Since the decision is made at the sequence level, all tokens follow the same path through the model, avoiding KV cache inconsistencies during decoding.Finally, we fine-tune the model with trained routers using LoRA adapters to recover any quality drop introduced by layer skipping, while preserving latency gains. LoRA fine-tuning also smoothens layer skipping
and further highlights the varied importance of layers based on input sequence. Our key contributions include:
\begin{enumerate}[noitemsep, nolistsep, leftmargin=*]
    \item We propose \textsc{FiRST}, a training and inference algorithm that uses layer-specific routers to perform input-adaptive layer selection. All tokens in a sequence follow the same selected layers, ensuring compatibility with KV caching and avoiding additional compute or latency overhead. \textsc{FiRST} is model-agnostic and can be applied on top of any pre-trained LLM.
    
    \item We introduce a LoRA-based fine-tuning approach on top of router-based layer selection to recover quality while maintaining latency gains. This also encourages smoother and a more stable layer selection.

    \item Finally, we conduct extensive experiments with \textsc{FiRST} across multiple datasets spanning three distinct tasks: Machine Translation, Summarization, and Question Answering. We evaluate on two open-source model architectures—LLaMA-3-8B and LLaMA-3.2-3B—and show that, for the same target speed-up, \textsc{FiRST} significantly improves performance across tasks as compared to baselines.

\end{enumerate}

\section{Problem Statement}

Our goal is to exploit the heterogeneity in inputs and tasks to selectively use LLM layers in a quality-aware manner for reducing inference latency and compute for on-device constraints. 
Ideally, we want to select an \emph{optimal} sub-sequence of layers within a transformer architecture for a given input and task, such that the overall latency, as well as expended computation, are both low, while quality is comparable to the un-modified case where every input sequence passes through every layer.
%for every task. 
For ease of explanation, without loss of generality, we assume the task is same and simply consider an input sequence for describing the problem. 

Let us consider an an input sequence $\mathcal{X} = \{x_1, x_2, \ldots, x_n\}$ with $n$ tokens. Let there be $m$ transformer layers in the model, where the $i^{th}$ transformer layer is represented as the function $\phi_i()$. 
As stated lucidly in \citep{llamas}, $\mathcal{X}$ is first converted to an initial latent representation $\mathcal{H}_0 = \{H^1_0, H^2_0, \ldots, H^n_0\}$, where $H^0_{j} \in \mathbb{R}^D, \forall j \in [n]$ is a look-up from a learned embedding dictionary corresponding to the $j^{th}$ token. Thereafter, every transformer layer 
$\phi_i()$ operates on the latent vectors $\mathcal{H}_{i}$ to generate the embedding for the $i^{th}$ layer as follows. For the $j^{th}$ token, the embedding at layer $i$ is computed as follows:
% for all the tokens in the sequence previous to $$ sequence from all the previous layer $k \in [i-1]$, to which the the residual from the previous layer is added to generate the resultant embedding after the $i^{th}$ layer $\mathcal{H}^i = \{H_^j_i\}, \ j \in [n]$, in the same dimension $D$ as follows \citep{llamas}: 
\begin{align}
\label{eq:autoregressive}
    H^j_i &= H^{j}_{i-1} + \phi_i(H^1_{i-1}, H^2_{i-1}, \ldots, H^j_{i-1})
\end{align}

Let the (gold) output or generated sequence for an input sequence $\mathcal{X}$ that passed through all $m$ layers of the model with full computation be $\mathcal{Y}^*_\mathcal{X}$. 
Our hypothesis is that for a given input sequence (and task), there exists an optimal subsequence of functions $\mathcal{F}_{OPT}(\mathcal{X})$ out of the full sequence $\{\phi_i, i \in [m]\}$ such that the output generated by passing through this subsequence:  $\mathcal{Y}_{OPT,\mathcal{X}} \approx \mathcal{Y}^*_\mathcal{X}$. 
More formally, if $Q$ is a quantitative quality measure on $\mathcal{Y}$, and $\epsilon \rightarrow 0$ is tolerance in deviation in quality from the gold output,  then 
we hypothesize that there exists an optimal subsequence, using the minimum number of layers, $\mathcal{F}_{OPT}(\mathcal{X})$, 
such that: 
% \todo[inline]{It is not immediately clear why the optimal subsequence cannot be the entire sequence. Need to add something here? I know it is explained in the line below but not clear from the equation.}
\begin{align}
Q\left(\mathcal{Y}_{OPT, \mathcal{X}}\right) &\geq (1-\epsilon) Q\left(\mathcal{Y}^*_\mathcal{X}\right), \forall \mathcal{X}.
\end{align}

The optimality above is with respect to the minimum subsequence of layers that can help achieve the above, to minimize latency while keeping quality unaffected. 
Note that, the optimal subsequence $\mathcal{F}_{OPT}(\mathcal{X})$ need to be obey the same autoregressive computation on previous tokens as given in Equation \ref{eq:autoregressive}. Hence, any algorithm that determines the optimal subsequence, need to be compatible with KV cache handling, to avoid the re-computation of values for tokens preceding the current token. %(which is a drawback with some existing work, especially in the Early Exit literature, that choose computation or layer skipping at token level). 

The potential number of subsequences for $m$ layers is $2^m$, hence a brute force is infeasible and also beats the purpose of such a layer selection in the first place: reducing latency and compute. In the absence of any known substructure in the behaviour of the latent layers on each input sequence, it is difficult to arrive at the optimal solution polynomially or with low additional latency or compute. %, and in fact is likely to be NP-hard. 

We propose to learn an approximation of the optimal subsequence of layers for any input sequence with low additional latency and minimal training. 

\section{Proposed Solution: \textsc{FiRST}}
\begin{figure}
    \centering    \includegraphics[width=0.3\textwidth]{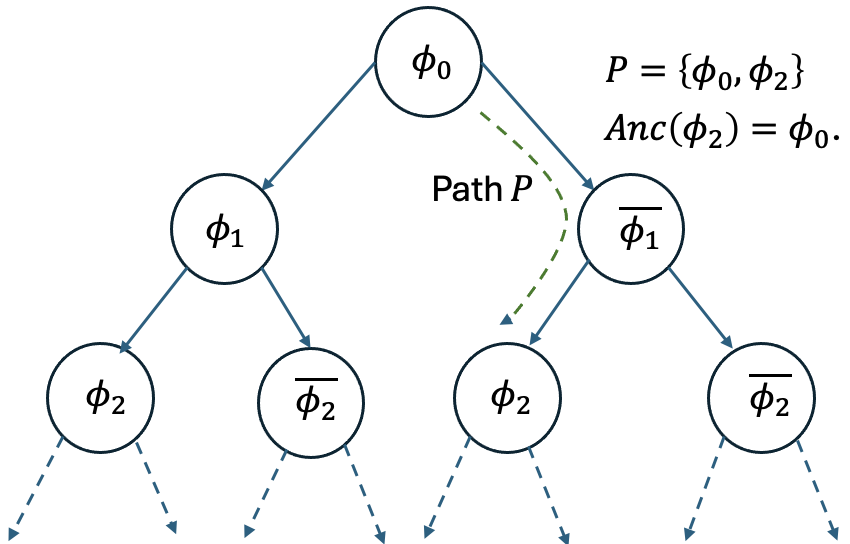}
    \caption{\small Binary Tree representation of layer selection.}
    \label{fig:binary}
\end{figure}

Let us first understand what it entails to learn an optimal subsequence of layers for any input. Consider the full transformer sequence to be $\mathcal{F}^* = \{\phi_1, \phi_2, \ldots, \phi_m\}$. Any optimal subsequence for an input $\mathcal{X}$: $\mathcal{F}_{OPT, \mathcal{X}}$ could be thought of as finding an optimal path through a binary tree of functions. Formally, let every level in the binary tree correspond to a transformer layer and the $0^{th}$ layer corresponds to the initial embedding look up; i.e., at depth $i \in [m]$, there would be $2^i$ nodes, each corresponding to either $\phi_{i}$ or $\overline{\phi_{i}}$, where the former denotes that a particular transformer layer is included in the optimal path whereas the latter denotes that it is not included.  
Each (of the $2^{i-1}$ nodes) $\phi_i$ or $\overline{\phi_i}$ has two children, corresponding to the next transformer layer: $\phi_{i+1}$ and $\overline{\phi_{i+1}}$ (See Figure \ref{fig:binary}). 
In such a tree structure, for example, the path 
$\{ \phi_{i}, \overline{\phi_{i+1}}, \phi_{i+2}\}$ indicates the subsequence of transformer layers $\{\phi_i, \phi_{i+2}\}$. 
For any transformer layer $\phi_{i}$ in this tree, let $Anc(\phi_i) = k, 0\leq k<i$ denote the the lowest ancestor node where the corresponding transformer node $\phi_k$ is included in the sequence. In the above example, $Anc(\phi_{i+2}) = \phi_{i}$.

Consider a sequence of functions $\mathcal{F}$, where for level $i$, $Anc(\phi_i) = \phi_k$. The autoregressive computations for the $j^{th}$ token in the input sequence (originally Eq \ref{eq:autoregressive}), would now be modified as: 
%\scriptsize
\begin{align}
\label{eq:modified}
H^j_i &= 
\begin{cases}
H^j_k, & \text{if } \phi_i \notin \mathcal{F}, \\
H^j_k + \phi_i(H^1_k, H^2_k, \ldots, H^j_k), & \text{if } \phi_i \in \mathcal{F}.
\end{cases}
\end{align}
%\normalsize
Our problem translates to navigating this binary tree to find the optimal path $\mathcal{F}_{OPT}$ for an input sequence and task. 
Since there are $2^m$ paths in this tree, we propose to approximate the optimal by making a decision in a greedy fashion at each node. Formally, we add a (lightweight and fast) router $R_i$ before every transformer layer $\phi_i$ in the model, that will predict whether $\phi_i$ will be selected or not. 

Our aim is to learn to predict the layer choice %
%We want to decide this 
at a sequence level (not token) to maintain compatibility with the autoregressive computations and avoid re-computation of of KV cache values. Moreover, we should  spend minimal compute %and training 
for learning the $R_i$ functions. Finally, $R_i$ functions should %be lightweight and low compute so that that do 
not add any significant latency to the overall computation.%, helping realize the latency gains. 

%Our proposed algorithm 
\textsc{FiRST} modifies any off-shelf pre-trained transformer based model by incorporating and training a router or probability function $R_i$ before every transformer layer $\phi_i$. 
%\textsc{FiRST} trains routers 
%During inference, %for a given input sequence $\mathcal{X}$, 
The output of $R_i$ is a %probability 
score $\rho_i$ denoting the probability of selecting $\phi_i$ in the layer sequence. 
During inference, $\rho_i$ is rounded to determine selection of $\phi_i$.  
%Let $\phi^R_i = \phi_i \text{ if } \rho_i \geq 0.5, \text{else } 0$. 
Let $\round{\rho_i} = 1 \text{ if } \rho_i \geq 0.5, \text{ else } 0$. Equation \ref{eq:autoregressive} is now modified as: 
%In other words, if $\rho_i\geq 0.5$, we select layer $i$, otherwise we skip it. %the la% + (1 - \rho_i)\cdot \phi_i(\mathcal{X})$. 
%The autoregressive computation in 
%The autoregressive computation in Eq \ref{eq:autoregressive} now becomes: 
% \begin{align}
% \label{eq:modified}
% H^j_i &= 
% \begin{cases}
% H^j_{i-1}, & \text{if } \rho_i < 0.5, \\
% H^j_{i-1} + \phi_i(H^1_{i-1}, H^2_{i-1}, \ldots, H^j_{i-1}), & \text{if } \rho_i \geq 0.5.
% \end{cases}
% \end{align}
% \begin{multline}
\begin{align*}
%\label{eq:first}
H^j_i &= H^j_{i-1} + \round{\rho_i} \cdot \phi_i(H^1_{i-1}, H^2_{i-1}, \ldots, H^j_{i-1}) 
\end{align*}
% \end{multline}

This recursively approximates Eq \ref{eq:modified} for the optimal $\mathcal{F}$ in a probabilistic, greedy manner. 
We train the functions $R_i$ on datasets and tasks, and further fine tune using LoRA adapters to make the layer selections smooth and improve the output quality. 
%We explain the framework for \textsc{FiRST} algorithm in details in the following section. 

\section{\textsc{FiRST} Framework and Algorithm}
In this section, we describe the training and inference frameworks for \textsc{FiRST} in details. We discuss how to train routers to be adaptive to input sequences. Given an off-the-shelf pre-trained LLM, we propose two training phases. In the first phase, we train a router for each layer that decides whether the input sequence should skip the layer. In the second phase, to tackle the issue of unseen skipping during pre-training, we fine-tune the router-augmented LLM keeping router weights fixed to ensure the model improves performance on the target dataset without reducing the skipping level. While joint training of the router and LoRA modules is theoretically possible, we find it introduces optimization instability (see Appendix~\ref{app:setup_appendix} for further discussion on this design choice and its impact). 
%In other words, the LoRA fine-tuning ensures that the gap in performance with and without skipping is significantly reduced when compared to the base model. Below, we provide the details of each phase.

% \begin{figure}
%     \centering
%     \includegraphics[width=\linewidth]{Images/our_solution_modules.png}
%     \caption{Layer diagram of the two training phases}
%     \label{fig:modules}
% \end{figure}

\subsection{Adaptive Router Module}

The adaptive router module is a single-layer neural network without bias, positioned before every layer in the model. During training of the router, all model parameters except the router weights remain frozen. For the first layer, it takes the tokenized input, and for each of the subsequent layers, it takes the output of the preceding layer as input. Mathematically speaking, for any layer $i$, given a batch of $B$ tokenized inputs sequences, where each sequence has $n$ tokens and is embedded in to $\mathbb{R}^D$, 
the adaptive router module  takes as input a $B\times n\times D$ tensor output of layer $(i-1)$ and outputs a $B\times n \times 1$ tensor. Subsequently, corresponding to each value (or, token) in the $B\times n\times 1$ tensor, we apply a sigmoid function to ensure that all entries in the tensor are in the interval $[0,1]$. Following this, we take a mean operation at the sequence level - we take a mean of all the weights in a sequence to output a $B\times 1\times 1$ tensor. For each sequence in the batch, the corresponding entry is the probability $\rho_i$ with which the sequence passes through the layer $i$. The input sequence skips the layer $i$ with probability $1-\rho_i$.  
%In summary, after applying router $R_i$ to an input sequence at each layer $i$, a single probability value $\rho_i$ is produced, indicating whether to pass the sequence through the layer. 
During training, the output of a layer is modified using a skip connection, incorporating the probability $\rho_i$ (see Figure: \ref{fig:layer_train_inference}). 

The routers are trained to encourage skipping by reducing the probabilities $\{\rho_i\}_i$ using a regularizer, to approximate the optimal subsequence for minimizing the latency. The training task is modeled as a language modeling task, specifically next token prediction.
The loss function comprises of 3 terms:  
\begin{itemize}[noitemsep, nolistsep, leftmargin=*]
    \item \textbf{Cross-entropy loss:} Standard difference between  actual and predicted probability distributions to ensure the quality of generation: 
    %\[
    $\mathcal{L}_{\mathrm{CE}} = - \sum_{x \in \mathcal{X}} \mathcal{Y}^*_\mathcal{X} \log(\hat{\mathcal{Y}})$.
    %\]
    \item \textbf{Regularization loss:} Adds a penalty term to reduce overfitting to noise: 
    $\mathcal{L}_{\mathrm{Reg}} = \sum_{i \in [m]} ||R_i||^2$, 
    where \(||R_i||^2\) denotes the \(\ell_2\) norm of the router weights for the \(i^{\text{th}}\) layer router, and there are \(m\) layers in the model. 
    \item \textbf{Non-skip penalization loss:} This is the summation of probability values across all layers of the model architecture: 
    %It encourages the model to favor skipping at the cost of cross-entropy loss to reduce latency, with the coefficient \(\alpha\) managing the extent of skipping to approximate the optimal trade-off between quality and latency. 
    $\mathcal{L}_{\mathrm{PP}} = \sum_{i \in [m]} \rho_i$
\end{itemize}
The total loss $\mathcal{L}$ is a linear combination of these three terms: $\mathcal{L} = \mathcal{L}_{\mathrm{CE}} + \lambda \cdot \mathcal{L}_{\mathrm{Reg}} + \alpha \cdot \mathcal{L}_{\mathrm{PP}}$, where \(\alpha\) manages the tradeoff between quality and latency. 
%manages the extent of skipping to approximate the optimal trade-off between quality and latency.
%\[
% \mathcal{L} = \mathcal{L}_{\mathrm{CE}} + \lambda \cdot \mathcal{L}_{\mathrm{Reg}} + \alpha \cdot \mathcal{L}_{\mathrm{PP}}
% \]

%Henceforth, we denote $\alpha$ as the \texttt{prob\_penalization} hyperparameter. 

% \begin{figure}
%     \centering
%     \includegraphics[width=0.25\linewidth]{Images/training.png}
%     \caption{Skip connection used for router training. With probability $p$, the sequence is processed by the layer and with probability $1-p$, the layer is skipped.}
%     \label{fig:training}
% \end{figure}

%Training is conducted until the combined loss remains stable for the next 5 batch gradient descent steps. During the training of the routers, the prompt includes only the task and instruction, without attaching the responses. This setup is designed to align with the inference environment, where the routers make decisions based solely on the initial sequence passed to them. 
% Note that we set the hyper-parameters during the training phase to be as follows:

% \begin{itemize}
%     \item The learning rate is set between $1e^{-4}$ and $3e^{-4}$.
%     \item Gradient accumulation steps are set to 4/5.
%     \item A cosine scheduler is used to adjust the learning rate during training.
%     \item The regularization loss coefficient $\lambda$ is fixed at 0.01.
%     \item The probability penalization coefficient $\alpha$ is tuned according to the dataset and sequence length. We show our results for a wide range of this hyper-parameter.
% \end{itemize}
\subsection{LoRA Compensation Module}

Skipping layers naturally leads to some performance loss - especially so since the pre-trained model was not trained to skip layers.
To compensate for the loss in performance caused by skipping layers, we finetune the router-augmented pre-trained model on the downstream task \footnote{similar to Quantization Aware Training such as QLoRA \cite{dettmers2024qlora} - compensates for model compression} using Low Rank Adapters (LoRA). % to modify the weights of the pre-trained model while keeping the router weights frozen.  
During finetuning, the router parameters are frozen while trainable LoRA adapters are added to both the FFN (Feed-Forward Network) and the attention modules of each layer of the pre-trained model. In order to maintain the skipping level, we again add a non-skip penalization loss component during finetuning with scaling hyper-parameter $\beta$. This is essential even though the router weights are frozen because standard finetuning alters the hidden representations of the input sequence in a manner such that no layers are skipped. Note that the LoRA adapters do not lead to any latency overhead during inference.
%We have noticed that the Non-skip Penalization Loss coefficient $\alpha$ scaled down by a factor of 3-4 is well-suited for the finetuning process while maintaining the same skipping level as in phase 1 of the training. 
%During training the LoRA adapters, responses are appended to the prompt to train the model to predict tokens from the response. 
%For inference, the model weights are merged with the original weights to prevent any latency overhead. 
%\vspace{-2mm}

\begin{table*}[t]
\centering
\scriptsize
\begin{tabular}{c|c|c|c|c|c|c |c |c|c|c}
\toprule
\multicolumn{1}{c|}{} & \multicolumn{2}{c|}{} & \multicolumn{4}{c|}{\textbf{En-to-De}} & \multicolumn{4}{c}{\textbf{En-to-Zh}} \\
\cmidrule{4-11}
\textbf{Skip (\%)} & \multicolumn{2}{c|}{\textbf{{Model Type}}} & \multicolumn{2}{c|}{\textbf{LLaMA-3-8B}}& \multicolumn{2}{c|}{\textbf{LLaMA-3.2-3B}} & \multicolumn{2}{c|}{\textbf{LLaMA-3-8B} }& \multicolumn{2}{c}{\textbf{LLaMA-3.2-3B}}\\[1.06ex]
% \cmidrule{4-9}
& \multicolumn{2}{c|}{}& \textbf{BLEU}& \textbf{COMET} & \textbf{BLEU}& \textbf{COMET}  & \textbf{BLEU}& \textbf{COMET} & \textbf{BLEU}&\textbf{COMET}  \\[1.06ex]
\midrule 
\multirow{2}{*}{\textbf{0}} & \multirow{2}{*}{Original Model} & Base + LoRA & \textbf{0.199}& \textbf{93.00}& \textbf{0.160}& \textbf{89.72}& \textbf{0.333}& \textbf{82.66}& \textbf{0.278}&\textbf{79.13}\\
 &  & Base & 0.169& 87.13& 0.125& 81.66& 0.208& 68.95& 0.166&61.84\\
\midrule
\multirow{8}{*}{\textbf{15}} & \multirow{2}{*}{Skip Decode} & Router + LoRA & 0.094& 55.62& 0.105& 44.58& 0.149& 55.98& 0.2&46.7\\
 &  & Router & 0.019& 23.33& 0.055& 32.74& 0.03& 21.75& 0.062&34.14\\
 & \multirow{2}{*}{Random Skip} & Router + LoRA & 0.097& 66.25& 0.079& 47.26& 0.237& 67.32& 0.154&57.79\\
 &  & Router & 0.132& 60.27& 0.048& 36.3& 0.168& 59.89& 0.077&35.86\\
 & \multirow{2}{*}{Unified Skip} & Router + LoRA & 0.153& 59.34& 0.095& 44.72& 0.23& \textbf{69.58}& 0.157&57.1\\
 &  & Router & 0.117& 59.26& 0.067& 39.81& 0.122& 54.57& 0.087&45.16\\
 & \multirow{2}{*}{\textbf{FiRST (Ours)}} & Router + LoRA & \textbf{0.161}& \textbf{82.14}& \textbf{0.113}& \textbf{60.29}& \textbf{0.247}& 68.63& \textbf{0.218}&\textbf{67.45}\\
 &  & Router & 0.108& 67.74& 0.069& 43.04& 0.08& 42.76& 0.1&54.55\\
\midrule
\multirow{8}{*}{\textbf{25}} & \multirow{2}{*}{Skip Decode} & Router + LoRA & 0.07& 31.47& \textbf{0.077}& 32.33& 0.088& 33.85& \textbf{0.147}&42.01\\
 &  & Router & 0.015& 21.55& 0.05& 27.64& 0.024& 20.93& 0.045&29.51\\
 & \multirow{2}{*}{Random Skip} & Router + LoRA & 0.018& 29.71& 0.023& 30.97& 0.065& 27.73& 0.137&\textbf{45.53}\\
 &  & Router & 0.011& 29.95& 0.015& 27.22& 0.04& 35.16& 0.053&31.5\\
 & \multirow{2}{*}{Unified Skip} & Router + LoRA & 0.06& 31.69& 0.05& 39.81& \textbf{0.142}& 50.59& 0.108&42.3\\
 &  & Router & 0.048& 32.15& 0.032& 30.86& 0.068& 38.74& 0.079&31.63\\
 & \multirow{2}{*}{\textbf{FiRST (Ours)}} & Router + LoRA & \textbf{0.071}& \textbf{34.95}& 0.072& \textbf{45.38}& 0.119& \textbf{56.92}& 0.126&41.66\\
 &  & Router & 0.029& 26.01& 0.029& 29.08& 0.051& 25.45& 0.053&27.83\\
\bottomrule
\end{tabular}
\caption{\small Machine Translation results for the English-to-German and English-to-Chinese task for both models: BLEU and COMET scores are reported across varying skip levels. \textbf{FiRST (Ours)} performs consistently well across both translation directions.}
\label{tab:wmt_reduced_combined}
\end{table*}

\subsection{Inference for \textsc{FiRST}}
\begin{figure}[t]
    \centering
    \begin{minipage}{0.2\textwidth}
        \centering
        \includegraphics[width=\linewidth]{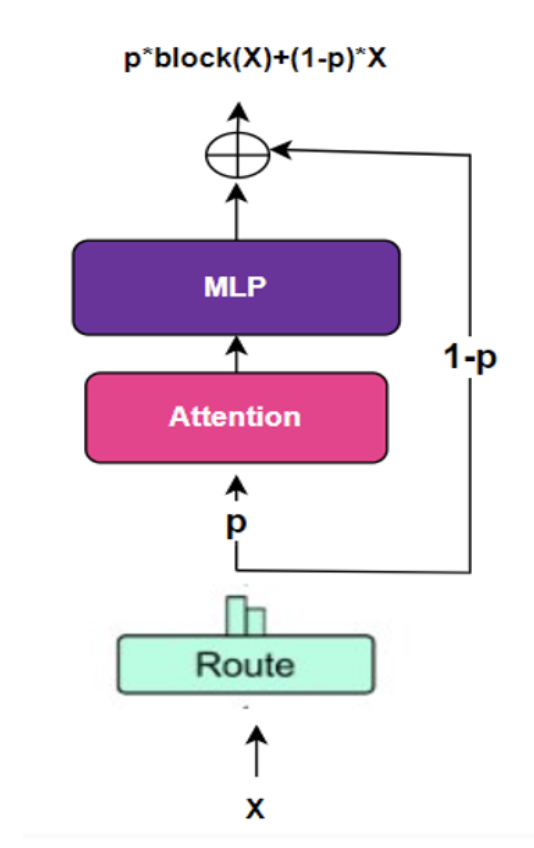}
    \end{minipage}
    \hspace{0.02\textwidth}
    \begin{minipage}{0.2\textwidth}
        \centering
        \includegraphics[width=\linewidth]{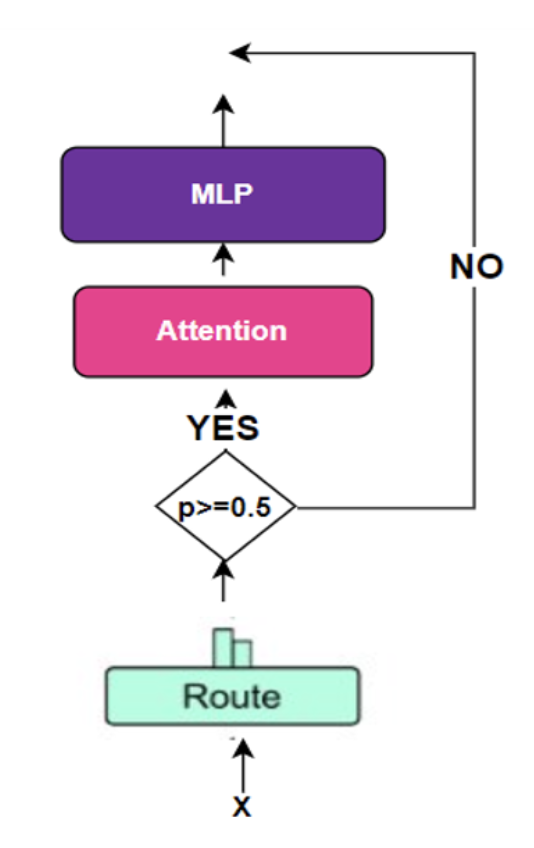}
    \end{minipage}
    \caption{\small Skip connection used for router training. With probability $p$, the sequence is processed by the layer and with probability $1-p$, the layer is skipped. During inference, routers make the decision of whether a sequence will skip a particular layer or pass through it.}
    \label{fig:layer_train_inference}
\end{figure}

During inference, for the input sequence, each router (corresponding to a layer) outputs a number in the interval $[0,1]$. If this number  is greater than or equal to 0.5, the sequence passes through the layer. Otherwise, the sequence skips the layer (Fig. \ref{fig:layer_train_inference}).
% \begin{figure}
%     \centering
%     \includegraphics[width=0.25\linewidth]{Images/inference.png}
%     \caption{During inference, routers make the decision of whether a sequence will skip a particular layer or pass through it }
%     \label{fig:inference}
% \end{figure}
%\newpage
Below, we discuss some salient points 
about the functioning of the router during inference to handle KV Cache appropriately:
\begin{enumerate}[nolistsep, noitemsep, leftmargin=*]
    \item \textbf{Prefill phase handling:} Skipping is not allowed during prefill phase. This ensures the first token is generated correctly, which is crucial for WMT tasks, as they are highly sensitive to the correct generation of the first token in the target language. It has been observed in prior works \citep{liu2024accelerating} that skipping during prefill phase is detrimental to performance during inference.
    \item \textbf{Fixed router decisions during decoding and handling KV Cache:} During the prefill phase, the decisions made by the routers are cached. During the decoding phase, every token adheres to the cached decision made during prefill. In other words, for a particular layer, if a router outputs a number less than 0.5 during prefill, the number is fixed for the decoding steps and therefore the same layer will be skipped by all tokens during decoding. Similarly, if the router outputs a number more than 0.5 during prefill, the same layer will be processing all tokens during decoding. Such a step ensures that for each decoding step and each layer that is not skipped, the KV cache for all previous tokens is available for that layer. This approach effectively addresses the caching issues encountered in early exit strategies, ensuring consistent decisions across the decoding process.
  %  -  this is because a fixed set of layers (decided during the prefill phase) will be skipped for all tokens during the decoding phase of inference.  
\end{enumerate}
% Hyper-parameters used during training and inference can be found in Appendix \ref{app:setup_appendix}.%have been included in Appendix 
\section{Experiments}

\begin{table}[t]
\centering
\scriptsize
\setlength{\tabcolsep}{4pt}
\begin{tabular}{ c |  c |  c  | c  | c | c }
\toprule
\textbf{Skip (\%)} & \multicolumn{2}{c|}{\textbf{Model Type}} & \multicolumn{2}{c|}{\textbf{SQuAD}} & \textbf{NQ} \\
 & \multicolumn{2}{c|}{} & \textbf{EM} & \textbf{F1} & \textbf{EM} \\
\midrule
\multicolumn{6}{c}{\textbf{LLaMA-3-8B}} \\
\midrule
\multirow{2}{*}{\textbf{0}} & \multirow{2}{*}{Original Model} & Base + LoRA & \textbf{73.93} & \textbf{85.99} & \textbf{51.88} \\
 &  & Base & 19.46 & 36.73 & 37.40 \\
\midrule
\multirow{2}{*}{\textbf{10}} & Skip Decode & R + LoRA & 60.14 & 65.33 & 41.98 \\
 &  & Router & 16.38 & 31.48 & 33.22 \\
\multirow{2}{*}{} & Random Skip & R + LoRA & 65.73 & 80.08 & 44.95 \\
 &  & Router & 18.25 & 33.75 & 34.09 \\
\multirow{2}{*}{} & Unified Skip & R + LoRA & 55.54 & 74.58 & 45.91 \\
 &  & Router & 17.39 & 32.91 & 33.64 \\
\multirow{2}{*}{} & \textbf{FiRST (Ours)} & R + LoRA & \textbf{70.85} & \textbf{83.61} & \textbf{47.85} \\
 &  & Router & 14.58 & 31.52 & 33.53 \\
\midrule
\multirow{2}{*}{\textbf{20}} & Skip Decode & R + LoRA & 45.00 & 55.10 & 26.62 \\
 &  & Router & 10.68 & 26.69 & 14.39 \\
\multirow{2}{*}{} & Random Skip & R + LoRA & 47.79 & 66.37 & 28.96 \\
 &  & Router & 6.71 & 22.46 & 27.40 \\
\multirow{2}{*}{} & Unified Skip & R + LoRA & 52.87 & 69.28 & 25.30 \\
 &  & Router & 18.18 & 32.51 & 23.57 \\
\multirow{2}{*}{} & \textbf{FiRST (Ours)} & R + LoRA & \textbf{60.60} & \textbf{75.49} & \textbf{32.22} \\
 &  & Router & 13.21 & 27.48 & 18.02 \\
\midrule
\multicolumn{6}{c}{\textbf{LLaMA-3.2-3B}} \\
\midrule
\multirow{2}{*}{\textbf{0}} & \multirow{2}{*}{Original Model} & Base + LoRA & \textbf{73.07} & \textbf{84.17} & \textbf{40.50} \\
 &  & Base & 18.92 & 37.74 & 30.10 \\
\midrule
\multirow{2}{*}{\textbf{10}} & Skip Decode & R + LoRA & 60.79 & 75.00 & 31.74 \\
 &  & Router & 20.00 & 31.55 & 21.51 \\
\multirow{2}{*}{} & Random Skip & R + LoRA & 64.78 & 77.27 & 36.02 \\
 &  & Router & 13.76 & 28.59 & 22.76 \\
\multirow{2}{*}{} & Unified Skip & R + LoRA & 65.03 & 77.53 & 32.90 \\
 &  & Router & 13.16 & 32.31 & 21.03 \\
\multirow{2}{*}{} & \textbf{FiRST (Ours)} & R + LoRA & \textbf{69.44} & \textbf{81.35} & \textbf{37.82} \\
 &  & Router & 12.79 & 28.37 & 22.55 \\
\midrule
\multirow{2}{*}{\textbf{20}} & Skip Decode & R + LoRA & \textbf{40.12} & 40.00 & 27.28 \\
 &  & Router & 20.45 & 37.62 & 14.73 \\
\multirow{2}{*}{} & Random Skip & R + LoRA & 11.32 & 38.34 & 26.60 \\
 &  & Router & 6.75 & 15.51 & 13.24 \\
\multirow{2}{*}{} & Unified Skip & R + LoRA & 37.39 & 52.49 & 26.84 \\
 &  & Router & 7.81 & 18.20 & 16.09 \\
\multirow{2}{*}{} & \textbf{FiRST (Ours)} & R + LoRA & 39.70 & \textbf{54.59} & \textbf{29.87} \\
 &  & Router & 5.52 & 15.33 & 17.51 \\
\bottomrule
\end{tabular}
\caption{\small{
Quality Analysis on Question Answering tasks using LLaMA-3-8B and LLaMA-3.2-3B across SQuAD (Exact Match and F1) and Natural Questions (Exact Match). Results are shown for multiple skip strategies and levels. \textbf{FiRST (Ours)} consistently performs best under skipping, demonstrating strong robustness. \textbf{R + LoRA} indicates Router Augmentation followed by LoRA fine-tuning.
}}
\label{tab:qa_combined}
\end{table}

\begin{table*}
\centering
\tiny
\begin{tabular}{ c | c | c | c | c | c }
\toprule
\textbf{Skip (\%)} & \multicolumn{2}{c|}{\textbf{Model Type}} & \textbf{BERT} & \textbf{R-1} & \textbf{R-L} \\
\midrule
\multicolumn{6}{c}{\textbf{LLaMA-3-8B}} \\
\midrule

% \hline
\multirow{2}{*}{0} & \multirow{2}{*}{Original Model} & Base + LoRA & \textbf{84.87} & \textbf{28.46} & \textbf{16.99} \\
 &  & Base & 82.29 & 23.49 & 14.66 \\
\midrule
\multirow{8}{*}{15} & \multirow{2}{*}{Skip Decode} & R + LoRA & 84.74 & 22.04 & 17.54 \\
 &  & Router & 82.53 & 13.68 & 9.30 \\
 & \multirow{2}{*}{Random Skip} & R + LoRA & 83.70 & 24.60 & 15.01 \\
 &  & Router & 81.10 & 19.64 & 13.07 \\
 & \multirow{2}{*}{Unified Skip} & R + LoRA & 84.25 & 24.35 & 14.3 \\
 &  & Router & 80.3 & 16.61 & 10.95 \\
 & \multirow{2}{*}{\textbf{FiRST (Ours)}} & R + LoRA & \textbf{85.14} & \textbf{31.8} & \textbf{20.13} \\
 &  & Router & 81.25 & 20.2 & 13.01 \\
\midrule
\multirow{8}{*}{20} & \multirow{2}{*}{Skip Decode} & R + LoRA & 82.57 & 20.41 & 14.87 \\
 &  & Router & 81.62 & 13.48 & 9.19 \\
 & \multirow{2}{*}{Random Skip} & R + LoRA & 81.39 & 21.57 & 13.83 \\
 &  & Router & 79.23 & 15.51 & 10.93 \\
 & \multirow{2}{*}{Unified Skip} & R + LoRA & \textbf{82.93} & 22.3 & 13.37 \\
 &  & Router & 80.32 & 16.51 & 11.15 \\
 & \multirow{2}{*}{\textbf{FiRST (Ours)}} & R + LoRA & 82.8 & \textbf{27.65} & \textbf{17.84} \\
 &  & Router & 79.32 & 16.28 & 10.85 \\
% \midrule
% \multirow{8}{*}{27} & \multirow{2}{*}{Skip Decode} & R + LoRA & 79.92 & 10.67 & 10.32 \\
%  &  & Router & 77.27 & 9.59 & 7.00 \\
%  & \multirow{2}{*}{Random Skip} & R + LoRA & 76.40 & 11.45 & 7.89 \\
%  &  & Router & \textbf{77.45} & 12.56 & 9.08 \\
%  & \multirow{2}{*}{Unified Skip} & R + LoRA & \textbf{80.28} & \textbf{15.94} & 9.89 \\
%  &  & Router & 77.43 & 10.97 & 7.68 \\
%  & \multirow{2}{*}{\textbf{FiRST (Ours)}} & R + LoRA & 77.5 & 14.65 & \textbf{10.45} \\
%  &  & Router & 75.6 & 9.39 & 6.92 \\
\bottomrule
\end{tabular}
\hspace{1cm}
\begin{tabular}{ c | c | c | c | c | c }
\toprule
\textbf{Skip (\%)} & \multicolumn{2}{c}{\textbf{Model Type}} & \textbf{BERT} & \textbf{R-1} & \textbf{R-L} \\
\midrule
\multicolumn{6}{c}{\textbf{LLaMA-3.2-3B}} \\
\midrule

% \hline
\multirow{2}{*}{0} & \multirow{2}{*}{Original Model} & Base + LoRA & \textbf{84.89} & \textbf{28.37} & \textbf{17.02} \\
 &  & Base & 71.85 & 19.34 & 12.00 \\
\midrule
\multirow{8}{*}{15} & \multirow{2}{*}{Skip Decode} & R + LoRA & \textbf{83.20} & 21.71 & 13.74 \\
 &  & Router & 80.97 & 9.74 & 6.87 \\
 & \multirow{2}{*}{Random Skip} & R + LoRA & 79.52 & 20.18 & 12.10 \\
 &  & Router & 68.20 & 10.10 & 7.10 \\
 & \multirow{2}{*}{Unified Skip} & R + LoRA & 81.53 & 18.89 & 11.72 \\
 &  & Router & 70.01 & 12.49 & 8.68 \\
 & \multirow{2}{*}{\textbf{FiRST (Ours)}} & R + LoRA & 83.17 & \textbf{26.47} & \textbf{16.79} \\
 &  & Router & 70.98 & 16.47 & 10.51 \\
\midrule
\multirow{8}{*}{24} & \multirow{2}{*}{Skip Decode} & R + LoRA & 78.55 & 15.83 & 6.74 \\
 &  & Router & 76.91 & 13.29 & 8.86 \\
 & \multirow{2}{*}{Random Skip} & R + LoRA & 80.00 & 16.33 & 10.07 \\
 &  & Router & 67.88 & 8.49 & 5.96 \\
 & \multirow{2}{*}{Unified Skip} & R + LoRA & 79.31 & 15.88 & 10.69 \\
 &  & Router & 68.86 & 9.17 & 6.97 \\
 & \multirow{2}{*}{\textbf{FiRST (Ours)}} & R + LoRA & \textbf{80.25} & \textbf{21.28} & \textbf{13.89} \\
 &  & Router & 69.17 & 12.36 & 8.43 \\
% \midrule
% \multirow{8}{*}{28} & \multirow{2}{*}{Skip Decode} & R + LoRA & 70.69 & 8.76 & 6.74 \\
%  &  & Router & 40.99 & 2.05 & 1.23 \\
%  & \multirow{2}{*}{Random Skip} & R + LoRA & \textbf{79.45} & 14.69 & 9.23 \\
%  &  & Router & 67.48 & 8.14 & 5.64 \\
%  & \multirow{2}{*}{Unified Skip} & R + LoRA & 78.57 & 11.74 & 7.47 \\
%  &  & Router & 68.23 & 8.12 & 5.66 \\
%  & \multirow{2}{*}{\textbf{FiRST (Ours)}} & R + LoRA & 77.48 & \textbf{15.98} & \textbf{11.14} \\
%  &  & Router & 67.14 & \textbf{8.09} & \textbf{6.00} \\
\bottomrule
\end{tabular}
\caption{\small Quality Analysis on Summarization (CNN/DM dataset) on LLaMA-3-8B (left) and LLaMA-3.2-3B (right): BERT F1, Rouge-1 and Rouge-L scores are reported for varying skipping levels. Note that R + LoRA corresponds to Router Augmentation followed by LoRA fine-tuning (in the proposed FiRST framework).}
\label{tab:cnn_dm}
\end{table*}

We conduct experiments on three benchmark tasks—Machine Translation, Text Summarization, and Question Answering—demonstrating the robustness and scalability of \textsc{FiRST} across diverse settings. We base our task selection on prior work in the field to ensure a fair and meaningful comparison \cite{liu2024accelerating,del2023skipdecode,schuster2022confidentadaptivelanguagemodeling}.

\noindent \textbf{Datasets:} For machine translation, we use WMT development sets (2017–2020) for English-to-Chinese and English-to-German tasks, evaluating performance on the WMT 2022 test set, which covers diverse domains such as news, social media, e-commerce, and conversational contexts. For summarization, we use the CNN/DailyMail dataset, with 4,000 randomly selected training samples and evaluation on the standard test set of 11,490 samples.
For question answering (QA), we utilize SQuAD v1.1 and Natural Questions (NQ). We train on 4,000 randomly selected samples from each dataset and evaluate on their respective validation sets as test set labels are unavailable. For NQ, we incorporate a retrieval step before answer generation, retrieving relevant passages as context. Appendix \ref{app:dataset_appendix} contains detailed descriptions.% of the datasets. %\ref{app:dataset_appendix}.}

\noindent \textbf{Evaluation Metrics:} We employ standard metrics to assess output quality across three distinct tasks. For Machine Translation, we benchmark performance using BLEU and COMET; COMET is included for its more nuanced assessment capabilities beyond the n-gram overlap measured by BLEU. Summarization quality is evaluated with ROUGE scores and BERT Score, the latter capturing meaning-based similarity. For Question Answering, we use Exact Match (EM) and F1 Score on the SQuAD dataset, and report Exact Match (EM) for Natural Questions (NQ). Finally, to benchmark latency, we measure Time Per Output Token (TPOT) on GPU, which gauges overall decoding performance. Detailed descriptions of all evaluation metrics are available in Appendix \ref{app:evaluation_metric_appendix}. Hyperparameters used during training and inference are provided in Appendix \ref{app:setup_appendix}.
%Finally, a detailed description of hyper-parameters used during training and inference have been included in Appendix \ref{app:setup_appendix}. 

\subsection{Baselines for comparison}
We report the latency improvement and quality numbers relative to the base models (no skipping). 

 \begin{itemize}[noitemsep, nolistsep, leftmargin=*]
 \item \textbf{Random Skipping:} We skip a set of $k$ layers randomly where $k$ depends on the target speedup. 
 \item \textbf{Skip Decode:}  We implement Skip Decode \cite{del2023skipdecode} method that features a monotonic decrease in processing layers, enabling later tokens to leverage the computational resources used for earlier ones. 
 \item \textbf{Unified Skipping: }This, to the best of our knowledge, is the state-of-the-art 
 method relies on using a heuristic-based strategy for retaining layers at fixed intervals. We replicate the algorithm in \citep{liu2024accelerating} and compare performance both with and without LoRA fine-tuning across various skipping percentages. 
\end{itemize}

    %For a particular target speedup ratio, the minimum and maximum exit layers are determined and the computation resources are allocated to each token from max to min in a linearly decaying fashion.}

\subsection{Experimental Results on Different Tasks}

% \textbf{Layer-wise Skipping Patterns:} First, we present layer-wise skipping statistics for the three tasks we experiment with. Notably, layer-wise skipping varies significantly across tasks, indicating that the importance of each layer is task- and dataset-dependent. For LLaMA-3-8B at a 15\% skipping rate, in English-to-German Machine Translation, layers 7–9 and 21 are fully skipped, while layer 18 is partially skipped. For English-to-Chinese, layers 7–9, 16, and 21 are fully skipped, with partial skipping in layer 20. In the Summarization task, layers 20, 22, and 23 are fully skipped, while layers 19, 21, and 26 are partially skipped. Some layers are skipped less than 10\% of the time, suggesting these layers are only necessary for specific sequences, highlighting the input-dependent and task-specific nature of layer importance. This is further demonstrated in the Question-Answering skipping plot, where only layer 22 is fully skipped, and different combinations of layers are skipped depending on the question. Detailed layer-wise skipping statistics can be found in Appendix \ref{app:skipping_stats}.
%Now, we present a detailed analysis of our experiments on WMT Translation, CNN Summarization and SQuAD datasets.

\noindent\textbf{WMT:} For LLaMA-3-8B, at $15\%$ skipping, \textsc{FiRST} achieves a latency improvement of upto 10\% on TPOT (see Tables \ref{tab:wmt_reduced_combined} and \ref{tab:tpot_results_main}).% while being competitive to the base model + LoRA (gold reference) in quality.
In most cases, it \textbf{significantly outperforms} other layer skipping strategies (Skip Decode, Random and Unified Skipping) and in other cases, it is comparable in quality. When compared to the gold output (Base + LoRA), \textsc{FiRST} generally retains a high percentage of the quality, for instance, often achieving $\geq 80\%$ of the COMET score for LLaMA-3-8B, and $\geq 70\%$ in BLEU scores. For $25\%$ skipping, \textsc{FiRST} achieves significant improvement in quality over other strategies, in almost all metrics, while achieving $\sim 18\%$ reduction in TPOT. %Results for English to Chinese translation can be found in Appendix: \ref{app:detailed_results_highlights}.

% For , applying approximately 15\% layer skipping leads to less than 20\% reduction in COMET scores (semantic metric), less than 15\% reduction in BLEU-1 scores, and less than 25\% reduction in BLEU-2 scores (syntax metrics) for both English-to-German and English-to-Chinese translations. This comes with a latency improvement of 10–12\% on TPOT (see Tables \ref{tab:wmt_reduced_8} and \ref{tab:tpot_total_8}). Our approach significantly outperforms the Unified Skipping baseline for English-to-German, achieving a BLEU-1 score of $38.01$ versus $28.92$ and a COMET score of $82.14$ versus $59.34$ for approximately $15\%$ skipping. For English-to-Chinese, performance remains competitive, with our method matching the baseline across BLEU and COMET metrics for $15\%$ skipping. 
\noindent For LLaMA-3.2-3B, for similar latency improvement ($\sim 10\%$), the COMET scores are significantly higher that other baselines while the BLEU scores are comparable. (Table \ref{tab:tpot_results_main}). It remains within $65 - 85\%$ of BLEU and COMET scores (Table \ref{tab:wmt_reduced_combined}) achieved by the gold standard.

\noindent \textbf{CNN/DailyMail Dataset:} For LLaMA-3-8B, at $\sim 15\%$ skipping, our method \textbf{outperforms} the Base + LoRA setting(Table: \ref{tab:cnn_dm}) while obtaining a 12\% improvement in TPOT (Table \ref{tab:tpot_results_main}).   
For LLaMA-3.2-3B, at $15\%$, the quality is comparable ($\sim 98\%$) to gold and other baselines with $12\%$ improvement in TPOT. At $24\%$, \textsc{FiRST} is significantly better than other layer skipping strategies, while achieving $>20\%$ improvement in latency. 
% the BERT-F1 scores are comparable to the baselines, there is over $\sim$ 30 \% and $\sim$ 20 \%improvement in ROUGE-1 and ROUGE-L over unified skipping for both skipping percentages with minimal degradation in performance over base model and 12-20 \% TPOT improvement.}

% \begin{table*}
% \renewcommand{\arraystretch}{1.2}
% \centering
% \scriptsize
% \begin{tabular}{ c  |c  |c |c }
% \toprule
% \textbf{Model Type} & \textbf{$\sim$Skipping (\%)} & \textbf{3-8B}&\textbf{3.2-3B}\\
% \textbf{} & \textbf{} & \textbf{TPOT}  &\textbf{TPOT}  \\
% \hline
% Base + LoRA & 0 & 1x  & 1x  \\
% \hline
% R + LoRA & 15 & 0.94x  & 0.94x  \\
% \hline
% R + LoRA & 24 & 0.83x  & 0.84x  \\
% \hline
% R + LoRA & 28 & 0.74x  & 0.72x  \\
% \bottomrule
% \end{tabular}
% \hspace{1cm}
% \begin{tabular}{ c  |c  |c | c }
% \toprule
% \textbf{Model Type} & \textbf{$\sim$Skipping (\%)} &  \textbf{3-8B}&\textbf{3.2-3B}\\
% \textbf{} & \textbf{} &  \textbf{TPOT}&\textbf{TPOT}  \\
% \hline
% Base + LoRA & 0&  1x&1x\\
% \hline
% R + LoRA & 10&  0.95x&0.96x\\
% \hline
% R + LoRA & 20&  0.78x&0.80x\\
% \bottomrule
% \end{tabular}
% \caption{\small{TPOT variation of LLaMA-3-8B (3-8B) and LLaMA-3.2-3B (3.2-3B) on SQuAD (left) and Natural Questions (right) dataset for FiRST. These values are relative to the LoRA fine-tuned base model. Fine-tuning improves TPOT and quality significantly.}}
% \label{tab:tpot_squad_nq}
% \end{table*}

\begin{table*}
\renewcommand{\arraystretch}{1.1}
\centering
\scriptsize
\begin{tabular}{c | c | 
                c c c | 
                c c | 
                c c | 
                c c}
\toprule
\textbf{Model size} & \textbf{Model Type} 
  & \multicolumn{3}{c|}{\textbf{WMT}} 
  & \multicolumn{2}{c|}{\textbf{CNN/DM}} 
  & \multicolumn{2}{c|}{\textbf{SQuAD}}
  & \multicolumn{2}{c}{\textbf{Natural Questions}} \\
\cmidrule(lr){3-5}\cmidrule(lr){6-7}\cmidrule(lr){8-9}\cmidrule(lr){10-11}
 & 
  & \textbf{Skip (\%)} & \textbf{Eng→De} & \textbf{Eng→Zh}
  & \textbf{Skip (\%)} & \textbf{TPOT}
  & \textbf{Skip (\%)} & \textbf{TPOT}
  & \textbf{Skip (\%)} & \textbf{TPOT} \\
\midrule
\multirow{4}{*}{LLaMA-3–8.B}
  & Base + LoRA &  0 & 1×   & 1×   &  0 & 1×   &  0 & 1×   &  0 & 1×   \\
  & R + LoRA    & 15 & 0.90×& 0.88×& 15 & 0.88×& 10 & 0.95×& 10 & 0.96×\\
  & R + LoRA    & 25 & 0.82×& 0.83×& 20 & 0.81×& 20 & 0.78×& 20 & 0.80×\\
\midrule
\multirow{4}{*}{LLaMA-3.2–3B}
  & Base + LoRA &  0 & 1×   & 1×   &  0 & 1×   &  0 & 1×   &  0 & 1×   \\
  & R + LoRA    & 15 & 0.90×& 0.91×& 15 & 0.88×& 15 & 0.94×& 15 & 0.94×\\
  & R + LoRA    & 25 & 0.78×& 0.75×& 24 & 0.79×& 24 & 0.83×& 24 & 0.84×\\
\bottomrule
\end{tabular}
\caption{\small TPOT variation across all datasets for FiRST. The reported values are relative to the LoRA fine-tuned base model. \textbf{R + LoRA} indicates Router Augmentation followed by LoRA fine-tuning. Fine-tuning improves TPOT and quality significantly.}
\label{tab:tpot_results_main}
\end{table*}

\noindent \textbf{SQuAD Dataset:}
For the LLaMA-3-8B model, \textsc{FiRST}(at 10\% skip level) maintains over 95\% of the performance of the gold standard (Base + LoRA without skipping)(Table \ref{tab:qa_combined}), with overall latency gains upto 16\% (Table \ref{tab:tpot_results_main}). It is \textbf{significantly} better in quality than all other baselines across all metrics for different levels of skipping. 
%Additionally, there is a 20\% increase in Exact Match over the unified skipping baseline at 10\% skipping, along with an 8-10\% improvement on other metrics for both skipping percentages. 
For LLaMA-3.2-3B, again \textsc{FiRST} is $> 95\%$ in output quality of gold (base + LoRA) for 10\% skipping (Table \ref{tab:qa_combined}) with similar gains in latency (upto 16\% overall) (Table \ref{tab:tpot_results_main}) over the LoRA fine-tuned base model. Moreover, it is better than all other layer skipping strategies across all metrics. 

\noindent \textbf{Natural Questions Dataset:}
For the LLaMA-3-8B model, \textsc{FiRST} retains over $92\%$ of the EM achieved by the non-skipping gold standard (Table \ref{tab:qa_combined}), with overall latency gains of 5-12\% (Table \ref{tab:tpot_results_main}). It is \textbf{significantly} better in quality than all other baselines for different levels of skipping. 
For LLaMA-3.2-3B, again \textsc{FiRST} is $> 93\%$ in output quality of gold (base + LoRA) for 10\% skipping (Table \ref{tab:qa_combined}) with gains in latency of 4-10\% overall (Table \ref{tab:tpot_results_main}) over the LoRA fine-tuned base model. Moreover, it is better than all other layer skipping strategies. While more sophisticated retrieval strategies could further enhance performance, our goal was to demonstrate that FiRST maintains quality performance and latency gains even in a retrieval-based QA setting.

\noindent Detailed results for an additional skipping percentage are provided in Appendix {\ref{app:detailed_results}}.

\begin{figure}
  \centering
\vspace{-4mm}
    \includegraphics[width=0.8\linewidth]{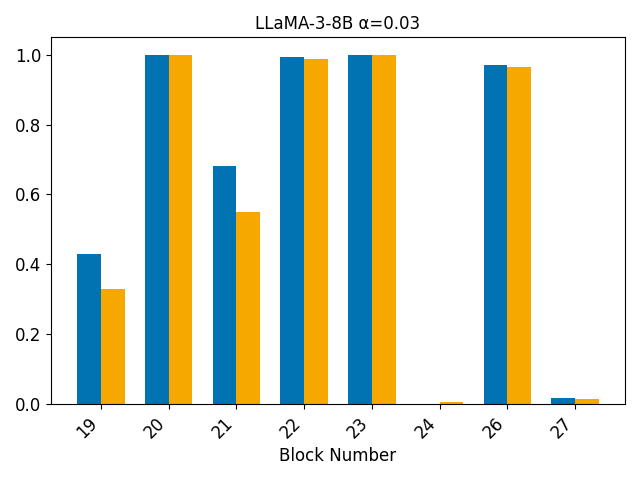}
  \caption{\small LLaMA-3-8B skipping statistics at 15\% skipping rate on summarization task. Layers with no skipping, indicated by a 0\% skipping rate, are not represented in the plot.}
\vspace{-4mm}
  \label{fig:skipping_cnn_main}
\end{figure}

\noindent \textbf{Layer-wise Skipping Patterns:} Layer-wise skipping varies significantly across tasks, reflecting the task-specific importance of each layer. For LLaMA-3-8B at a 15\% skipping rate, layers 7–9 and 21 are fully skipped in English-to-German, with partial skipping in layer 18. Figure~\ref{fig:skipping_cnn_main} for summarization shows that only some layers in the middle of the network, specifically layers 19 to 27, are skipped. The early layers and the last few layers (28–32) are never skipped.  Layers 20, 22, and 23 are fully skipped, with partial skipping in layers 19 and 21. Some layers are skipped less than 10\% of times, indicating their necessity for specific sequences.  This shows that it’s important to learn which layers to skip based on each input, as the skipping pattern is not the same for every sample. The task-specificity is also evident in Question-Answering, where only layer 22 is fully skipped on SQuAD while layers 12, 23 are fully skipped in case of NQ dataset, and skipping patterns depend on the input.
Detailed statistics are in Appendix \ref{app:skipping_stats}. 
 Furthermore, the router is trained in a task-aware manner, ensuring that skipping decisions align with task complexity. 
%Our results demonstrate that these trained routers generalize effectively (see Appendix: \ref{app:generalizability and Reusability of Routers}. 
We also conduct experiments to validate the reusability and generalizability of routers (see Appendix: \ref{app:generalizability and Reusability of Routers}) across different datasets for the same underlying task. 

\noindent \textbf{Computational overhead of Routers:}
We aim to improve the efficiency of pre-trained models by introducing routers, lightweight linear classifiers that help decide whether to skip certain layers during inference. A common concern with adding such components is the potential increase in computational cost. However, our analysis shows that the additional parameters and operations introduced by the routers account for a very small fraction of the total model computation, only 0.0027\% for LLaMA-3-8B and 0.0016\% for LLaMA-3.2-3B in terms of parameter count, and around 0.3\% of the total forward pass time. This fixed cost remains negligible compared to the overall model execution time. As a result, the overall latency gain primarily depends on how much computation is skipped, rather than the routing mechanism itself. Given similar levels of skipping, different methods tend to show comparable efficiency gains, so the key focus becomes minimizing the performance drop relative to the original model. Our results show that the proposed routing mechanism achieves this balance effectively, preserving latency improvements while maintaining model quality.

\section{Conclusion}
We propose a new framework, \textsc{FiRST}, for layer selection that adapts to the input sequence and task, aiming to reduce latency in a quality-aware manner. This approach is sequence-dependent and operates compatibly with KV caching.
With an optimal layer skipping rate of around 15\%, \textsc{FiRST} achieves a 10-20\% reduction in latency. This speedup is achieved while remaining quality-neutral, maintaining approximately 80\% or more of the base model's performance on quality metrics across multiple tasks (Machine Translation, Summarization, Question Answering) and model architectures on well-known open-source datasets. Furthermore, our method significantly outperforms other layer selection strategies on most quality metrics.

\newpage

\section{Limitations}
\textsc{FiRST} algorithm selects layers in a greedy, myopic way one layer at a time, corresponding to sequences (and tasks). A more optimal way of doing this would be to estimate a subsequence of layers to traverse through instead of one layer at a time. We intend to address this in future work. We would like to select a more optimal subset (subsequence) of layers which will increase the output quality while reducing latency even further.  
\section{Ethical Concerns}
There are no ethical concerns to the best of our knowledge.

\bibliography{ref}
\bibliographystyle{iclr2025_conference}

\newpage

\appendix

\section{Appendix}

\subsection{Related Work}
\label{app:related_work_appendix}

\paragraph{Early Exit:} Several works have been proposed in the early exit theme \citep{zhu2021leebert,zhou2020bert,xin2020deebert,liu2020fastbert,li2020cascadebert,hou2020dynabert,schuster2022confident,wu2020dontreaditadaptive} where adaptive compute is used for different parts of the token sequence. While these approaches have been popular for encoder-only models which processes the entire sequence as a whole, they have faced challenges in generation tasks. The main limitation of these set of techniques are their inability to handle KV caching appropriately which is crucial for multi-fold speed-ups in current LLM architectures. We emphasize that in our work, we assign varying compute to sequences in different batches but within the same sequence, we assign the same compute to every token. 

\paragraph{Input Agnostic Heuristics:} 
In Skip Decoding \citep{del2023skipdecode}, initial tokens pass through more layers than later ones, contradicting the observation that later tokens are harder to decode \citep{liu2024accelerating}. Additionally, Skip Decoding skips several bottom layers for most tokens, causing undesirable sub-network imbalance. To address this, Unified Layer Skipping \citep{liu2024accelerating} proposes a discrete skipping strategy that is uniform for all tokens in a sequence. Based on a latency budget, retained layer ids are passed through by all tokens, ensuring KV Cache handling and retaining key layers. However, the limitation of this approach is that skipping is independent of the input sequence. In contrast, early exit strategies adapt layer skipping to the input sequence, offering more flexibility. In \citep{fan2019reducing}, a method akin to dropout randomly skips layers during training, but this leads to performance decline during the pre-fill stage. FFN-SkipLLM \citep{jaiswal2024ffn} constrains skipping to FFN layers to avoid KV Cache issues but fails to fully exploit redundancy as discussed already. \cite{song2024sleb} is a very recent work that also explores greedily identifying layers to skip while preserving the model performance on a calibration dataset - however there are two major limitations of this work which are resolved in our paper - (A) first of all, the layer selection strategy is sequence independent although it can be made task-dependent by calibrating on task-specific datapoints - our approach for skipping layers is sequence dependent and is based on the input to a layer (B) SLEB does not explore the impact of fine-tuning on the layers to be skipped. On the other hand, our skipping strategy incorporates the trained router already - intuitively, the knowledge of skipping is transferred during finetuning. \cite{chen2024compressing} is another recent work that compresses models by identifying redundant layers - this is done by computing the average similarity between input/output pairs of a layer. However, as outlined in \cite{song2024sleb}, such an approach suffers from the limitation that it does not take into account the joint association between the layers while skipping multiple layers. Moreover, like \citep{jaiswal2024ffn}, this work is neither sequence dependent nor takes the final model predictions into account while identifying the layers to skip. 

\paragraph{Model Compression and Quantization Aware Training:} 
Orthogonal approaches to explore the latency/memory-performance trade-off in Large Language Models aim to build smaller models that approximate the performance of larger ones with reduced memory and latency costs. Key techniques include: 1) compressing model parameters into fewer bits \citep{frantar2022gptq,lin2024awq,lee2024owq,saha2023matrix}; 2) pruning the network by removing components like attention heads or neurons based on heuristics \citep{frantar2023sparsegpt,ma2023llm}; and 3) distilling the large model into a smaller, faster counterpart \citep{agarwal2023gkd,gu2024minillm}. For further details, we refer to the survey by \citep{zhu2023survey}. A significant body of work \citep{dettmers2024qlora,liu2023llm,peri2020deploying,li2023loftq} has focused on quantization-aware training to reduce memory footprints and mitigate performance loss, starting with QLoRA \citep{dettmers2024qlora}. In a similar vein, our work proposes fine-tuning router-augmented models to improve layer skipping and reduce performance degradation, as pre-trained models do not account for layer skipping, leading to higher degradation with vanilla skipping.

\paragraph{Network Pruning:} Another orthogonal approach to improve the inference speed-up is to prune redundant network weights by zeroing them out. There has been a significant body of work on pruning model weights \cite{frantar2022gptq, frantar2023sparsegpt, sun2022simple, zhang2023dynamic}  - most of these works can be categorized into two clusters namely unstructured pruning and structured pruning. In case of unstructured pruning, there is no structure to the inserted zeros and achieving speedups with modern GPU hardware tailored towards dense matrix multiplication is challenging. In fact, more than 90\% sparsity is typically required to achieve any significant speedup \cite{wang2020sparsert, shi2020efficient}. Therefore, structured pruning which is more amenable to GPU hardware has become prominent (2:4 pruning and sub-channel pruning). However, realizing desired speedups through these techniques have been difficult \cite{song2024sleb}. Moreover, several approaches for dynamically deleting entire rows or columns of weight matrices have been proposed \cite{ma2023structural, ashkboos2024slicegpt, liu2023deja} to retain dense matrices but two limitations remain - (A) hardware support is extremely limited for realizing speedup gains (B) extensive finetuning is necessary to align the sparsification with linguistic abilities - this is because, such pruning techniques were not observed by the model during pre-training. Finally, note that several prior works \cite{tang2024quest, oren2024transformers, xiao2023efficient} have imposed (query aware/ query agnostic) sparsity in the KV cache matrices to speed up self-attention mechanism via clever selection of the critical tokens necessary from the KV cache.
\paragraph{Mixture of Experts} Mixture of Experts (MoE) is a well-established technique for improving the efficiency and capacity of deep learning models by conditionally activating subsets of parameters for different inputs. MoE-based transformer models, such as Switch Transformers (Fedus et al., 2021) and GShard (Lepikhin et al., 2020), employ a gating mechanism to route tokens to a subset of expert layers, thereby significantly reducing computational costs while maintaining expressivity. These architectures are designed to scale up model capacity without a proportional increase in inference cost. The main goal is to scale up the parameters while maintaining the cost of pre-training and inference.

Although MoE and input-adaptive layer skipping share the goal of selective computation, they differ in fundamental ways. MoE dynamically routes tokens to different experts at the layer level, whereas layer skipping focuses on bypassing entire layers in the transformer stack based on the input query. MoE models typically maintain a full-depth model structure, leveraging sparse activation to reduce computational overhead, whereas layer-skipping methods explicitly modify the depth of computation for different inputs. Simply put, the techniques are orthogonal - layer skipping can be used in tandem with Moe to further reduce the depth of computation wherever possible at an expert level

\subsection{Details of Datasets}
\label{app:dataset_appendix}

\begin{table}[h]
\centering
\tiny
\begin{tabular}{l|cc|c|cc}
\toprule
 & \multicolumn{2}{c|}{\textbf{WMT}} & \textbf{Summarization} & \multicolumn{2}{c}{\textbf{Question Answering}} \\
\cmidrule(r){2-6} 
\textbf{Split} & \textbf{En$\rightarrow$De} & \textbf{En$\rightarrow$Zh} & \textbf{CNN/DM} & \textbf{SQuAD} & \textbf{NQ} \\
\midrule
Train        & 3,505 & 8,983 & 3,400  & 3,400  & 3,400 \\
Validation   &   876 &   998 &   600  &   600  &   600 \\
Test         & 2,038 & 2,038 & 11,490 & 10,570 & 7,830 \\
\bottomrule
\end{tabular}
\caption{Train, validation, and test splits for machine translation (WMT), summarization (CNN/DM), and question answering (SQuAD, NQ) tasks.}
\label{tab:train_test_split}
\end{table}

\textbf{Machine Translation:} For translation tasks, namely English-to-Chinese and English-to-German, we employ the WMT development sets from 2017 to 2020 for training/fine-tuning following the methodology outlined in previous studies \citep{liu2023instructionpositionmatterssequence, jiao2023parrottranslatingchatusing}. Translation performance is evaluated using the test set from the WMT 2022 dataset \citep{Kocmi_Bawden_Bojar_Dvorkovich_Federmann_Fishel_Gowda_Graham_Grundkiewicz_Haddow_et_al._2022} which was developed using recent content from diverse domains. These domains include news, social media, e-commerce, and conversational contexts.
\newline
\textbf{Summarization:} We use the popular CNN-DailyMail (CNN/DM) \citep{NIPS2015_afdec700} dataset which is a large collection (over 300k) of text summarization pairs, created from CNN and Daily Mail news articles. Each datapoint in this dataset comprises of an article (the body of the news article with 683 words on average) and the corresponding  highlights (article summary as written by the article author).
While the training set contains more than 287k samples, we have randomly chosen 4k samples for training both routers and LoRA. During training in our framework, the number of trainable parameters is small in both phases - therefore a small subset of data points is sufficient for training. 

\textbf{Question Answering:} We use the popular \textbf{Stanford Question Answering Dataset (SQuAD v1.1) }\citep{DBLP:journals/corr/RajpurkarZLL16}, a widely-used benchmark for Machine Question Answering. The dataset consists of over 100k question-answer pairs posed by crowd-workers on a set of over 500 Wikipedia articles. Each sample comprises a context (a passage from a Wikipedia article), a question (crafted to test comprehension of the passage), and the corresponding answer(a text span from the corresponding reading passage). Similarly to the CNN/DM dataset, 4k samples are chosen at random to train both routers and LoRA. The training and validation splits contain 87,599 and 10,570 samples, respectively. Evaluation is performed on the validation set \cite{schuster2022confident} as the test set labels are not publicly released. \textbf{Natural Questions (NQ):} We use the Natural Questions dataset \citep{kwiatkowski-etal-2019-natural}, which consists of real queries issued to Google Search paired with relevant Wikipedia articles. Each example in NQ contains a query (an actual user question), a context(Wikipedia article), and two types of answers: a long answer (typically a paragraph) and a short answer (a specific text span, when available). We implement a naïve RAG solution using a simple S-BERT model to generate embeddings for the query and passages and retrieve the top 5 most relevant passages based on similarity. Once retrieved, these passages are used as context to answer the query and then compared against gold answers. Similar to SQuAD, 4k samples are randomly chosen to train both routers and LoRA. The training set contains over 300k examples, with evaluation performed on 7.83k validation samples containing short answers. For further details on training-testing split, refer to table: {\ref{tab:train_test_split} 

\subsection{Evaluation Metrics}
\label{app:evaluation_metric_appendix}
\textbf{Quality-Based Metrics for Translation task:}
\begin{itemize}[noitemsep, nolistsep, leftmargin=*]
        \item \textbf{BLEU Score:} BLEU (Bilingual Evaluation Understudy) scores are used to measure the quality of translations. BLEU compares n-grams of the candidate translation to n-grams of the reference translation, providing a score between 0 and 1, with higher scores indicating better translations. In this evaluation, NLTK BLEU is employed and We report BLEU-1, BLEU-2, and the cumulative BLEU score, which is computed as the geometric mean of individual n-gram precision scores from unigram to 4-gram. To mitigate the issue of zero counts for higher-order n-grams, we apply the smoothing strategy utilized in \cite{post-2018-call}.
        
        \item \textbf{COMET:} COMET (Cross-lingual Optimized Metric for Evaluation of Translation) is used to assess translation quality further. COMET evaluates translations using a model trained to correlate well with human judgments. Specifically, Unbabel/XCOMET-XL \footnote{https://github.com/Unbabel/COMET} is used in this evaluation. COMET provides a more nuanced assessment of translation quality by considering the intricacies of both source and target languages, beyond the n-gram matching used in BLEU.
    \end{itemize}
  % \saransh{We report BLEU-1 and BLEU-2 scores, along with the cumulative BLEU score that accounts for individual precision scores from unigrams to 4-grams. BLEU scores serve as n-gram-based quality metrics. In addition, we include COMET, which offers a more robust semantic evaluation that goes beyond simple n-gram overlap.}
    
\textbf{Quality based Metrics for Summarization Task:} 
\begin{itemize}[noitemsep, nolistsep, leftmargin=*]
     \item \textbf{BERTScore:} This metric quantifies semantic similarity between texts by leveraging contextual word embeddings.
     
     BERTScore captures meaning-based similarity rather than relying on exact word matches, providing a nuanced evaluation of text generation quality.
        \item \textbf{ROUGE:}  (Recall-Oriented Understudy for Gisting Evaluation) is a common metric - ROUGE-1 refers to overlap of unigrams between the system summary and reference summary. Similarly, ROUGE-L measures longest matching sequence of words. 
        
    \end{itemize}
\AJ{\textbf{Quality based Metrics for Question Answering Task:} 
\begin{itemize}[noitemsep, nolistsep, leftmargin=*]
     \item \textbf{Exact Match:} This metric measures the percentage of predictions that exactly match the ground truth answer.
    \item \textbf{F1 score:} Since EM is a highly stringent metric, we also report the F1 score which provides a more flexible evaluation of answer prediction. This metric also takes into account near-matches.
\end{itemize}}

\subsection{Training and Inference Setup}
\label{app:setup_appendix}

\begin{itemize}[noitemsep, nolistsep, leftmargin=*]
\item \textbf{Training settings:} We perform extensive experiments on two models, namely LLaMA-3-8B and LLaMA-3.2-3B from Meta, which consist of 32 and 28 layers, respectively. Training of routers and LoRA adapters is conducted on A100 80GB GPUs, with training/inference is performed in full precision to avoid performance degradation due to quantization. The training process employs our custom loss function and continues for a fixed number of epochs, terminating when the validation loss fails to improve over 4 consecutive steps. The learning rate is set between $1e^{-4}$ and $3e^{-4}$ - a cosine scheduler is used to adjust the learning rate. Gradients are accumulated after five steps, and we set the regularization loss coefficient ($\lambda = 0.01$), ensuring it meaningfully contributes to the overall loss without overpowering primary losses like cross-entropy and penalization loss. After training, we verify the router weight norms (e.g., thresholds of $0.1$ or $0.05$) to ensure they remain stable—neither exploding nor vanishing—preventing overfitting or underfitting. For LoRA fine-tuning, we employ a rank of 8, a dropout rate of 0.1, and a scaling factor ($\text{lora\_alpha} = 32$). Our approach involves two key penalization coefficients: $\alpha$ for the router and $\beta$ for LoRA training. The non-skip coefficient ($\alpha$) is adjusted based on dataset characteristics, sequence length, and model depth, requiring some tuning. Typically, router penalization ($\alpha$) is set 2–4× higher than LoRA’s ($\beta$) based on experimental observations to maintain a consistent skipping percentage across both training phases. A reasonable starting point is to tune $\alpha$ for a 10–15\% skipping range, then gradually increase it at regular intervals to encourage higher skipping levels. In addition to the above hyperparameters, we define the maximum sequence length based on the task. For translation, it is set to 128 for router training and 256 for LoRA training. Similarly, for summarization, the sequence length is set to 500 and 700, respectively.
For Question Answering, this length is set to 512. The prompts for the different tasks related to training / inference are shown in Appendix \ref{app:prompt_details}.
\item \textbf{Joint training considerations:} The decision to split training into two phases—first training the router and then the LoRA modules—was made to ensure better stability during optimization. In our experiments, this approach allowed the router to establish a robust skipping strategy without interference from the fine-tuning of LoRA modules, which otherwise introduces instability. While training both the router and LoRA modules together is theoretically possible and could eliminate the need for reapplying the non-skip penalty in the second phase, initial experiments revealed that it led to suboptimal convergence and conflicting gradients. This was particularly evident when the router’s decisions were not yet stable, causing inconsistencies in the joint optimization process.
    \item \textbf{Inference settings:} \AJ{ For all the tasks, we set the temperature to 0.8 and enable top-k sampling over 10 tokens. The maximum number of tokens to be generated is set to 80 for WMT, 200 for CNN/DM and 32 for SQuAD and NQ.} Caching is turned on during inference.
\end{itemize}
% For summarization and translation tasks, we set the temperature to 0.8, with a beam size of 1 and top-k sampling over 10 tokens. For the question-answering tasks, the temperature is reduced to 0.2. During training, we perform a sweep over learning rates in the range of 3e-5 to 1e-4.
% We show comparison to the base model (Llama-3-8B) both with and without LoRA fine-tuning. While reporting TPOT, we assign the LoRA fine-tuned model (with no skipping) a baseline speedup of 1x. We then vary the coefficient $\alpha$ of the non-skip penalization loss to measure the corresponding performance (latency) improvements and resultant quality. Speedup measurements are reported relative to the base model, which natively supports key-value (KV) caching. 

\begin{figure*}[t]
  \centering
  \begin{minipage}{0.48\linewidth}
    \centering
    \includegraphics[width=\linewidth]{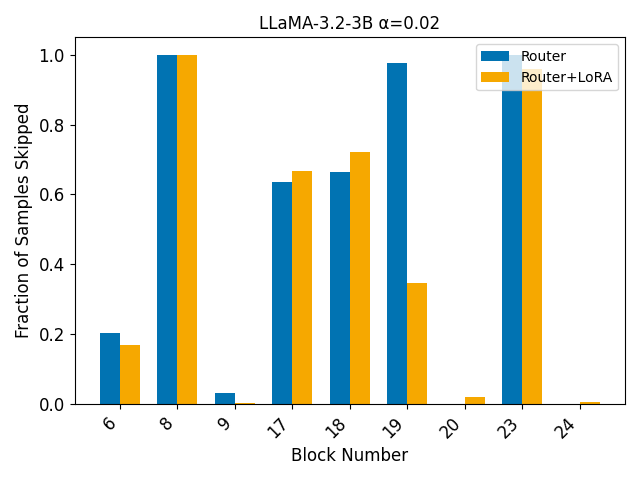}
  \end{minipage}%
  \hspace{0.05cm}  % Horizontal space between the top images
  \begin{minipage}{0.48\linewidth}
    \centering
    \includegraphics[width=\linewidth]{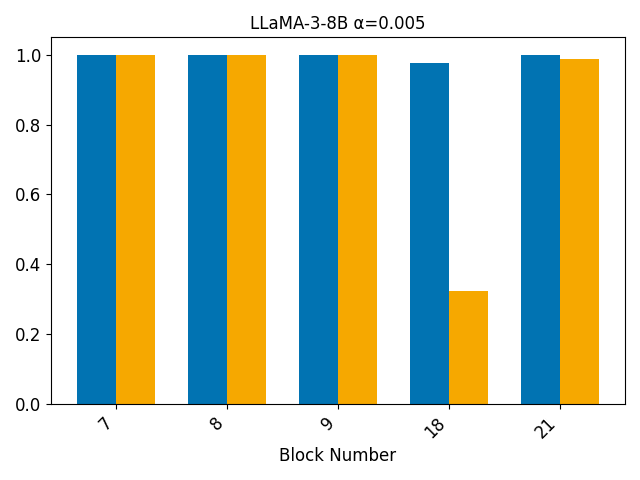}
  \end{minipage}
  \caption{Comparison of LLaMA-3.2-3B (left) and LLaMA-3-8B (right) at 15\% skipping rate on English-to-German Machine Translation Task. The graph shows how different layers contribute to the skipping behavior for the same dataset. Layers with no skipping, indicated by a 0\% skipping rate, are not represented in the plot.}
  \label{fig:skipping_eng_de_15}
\end{figure*}

\begin{figure*}[t]
  \centering
  \begin{minipage}{0.48\linewidth}
    \centering
    \includegraphics[width=\linewidth]{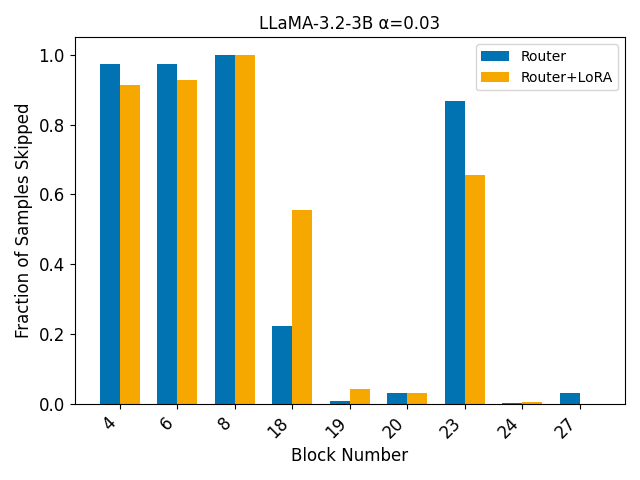}
  \end{minipage}%
  \hspace{0.05cm}  % Horizontal space between the top images
  \begin{minipage}{0.48\linewidth}
    \centering
    \includegraphics[width=\linewidth]{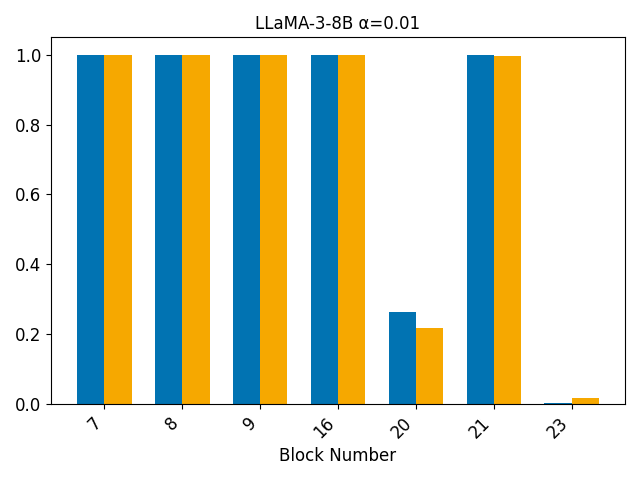}
  \end{minipage}
  \caption{Comparison of LLaMA-3.2-3B (left) and LLaMA-3-8B (right) at 15\% skipping rate on English-to-Chinese Machine Translation Task. The graph shows how different layers contribute to the skipping behavior for the same dataset. Layers with no skipping, indicated by a 0\% skipping rate, are not represented in the plot.}
  \label{fig:skipping_eng_zh_15}
\end{figure*}

\begin{figure*}
  \centering
  \begin{minipage}{0.48\linewidth}
    \centering
    \includegraphics[width=\linewidth]{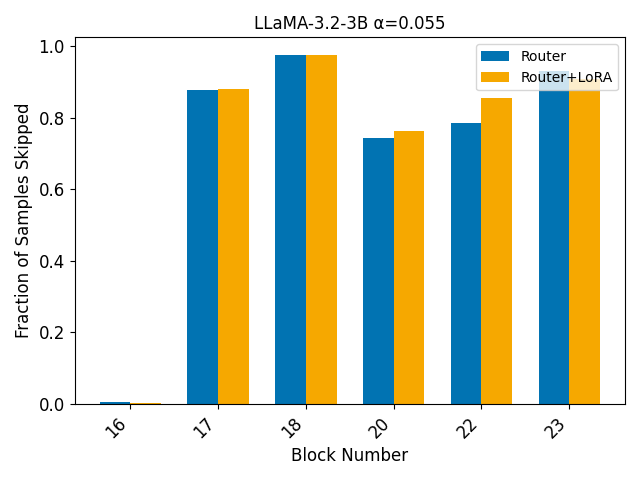}
  \end{minipage}%
  \hspace{0.05cm}  % Horizontal space between the top images
  \begin{minipage}{0.48\linewidth}
    \centering
    \includegraphics[width=\linewidth]{Images/cnn_15_8.png}
  \end{minipage}
  \caption{Comparison of LLaMA-3.2-3B (left) and LLaMA-3-8B (right) at 15\% skipping rate on CNN Summarization Task. Layers with no skipping, indicated by a 0\% skipping rate, are not represented in the plot.}
  \label{fig:skipping_cnn_15}
\end{figure*}

\begin{figure*}
  \centering
  \begin{minipage}{0.48\linewidth}
    \centering
    \includegraphics[width=\linewidth]{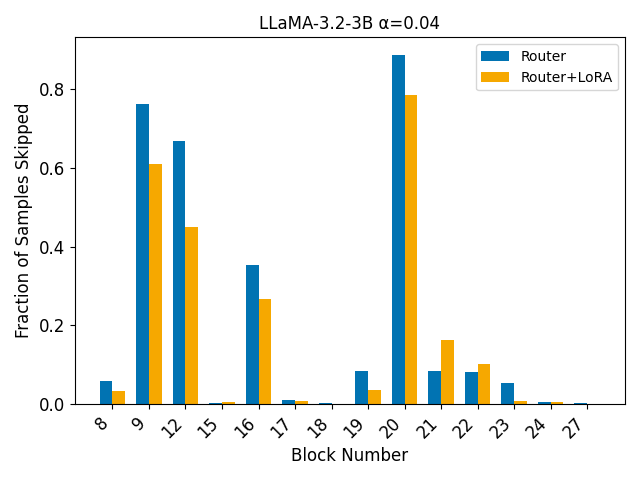}
  \end{minipage}%
  \hspace{0.05cm}  % Horizontal space between the top images
  \begin{minipage}{0.48\linewidth}
    \centering
    \includegraphics[width=\linewidth]{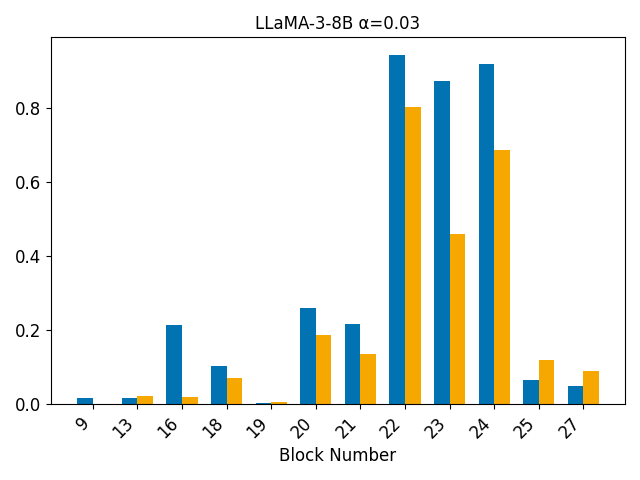}
  \end{minipage}
  \caption{Comparison of LLaMA-3.2-3B (left) and LLaMA-3-8B (right) at 15\% skipping rate on SQuAD Question-Answering Task. Layers with no skipping, indicated by a 0\% skipping rate, are not represented in the plot.}
  \label{fig:skipping_squad_15}
\end{figure*}

\begin{figure*}
  \centering
  \begin{minipage}{0.48\linewidth}
    \centering
    \includegraphics[width=\linewidth]{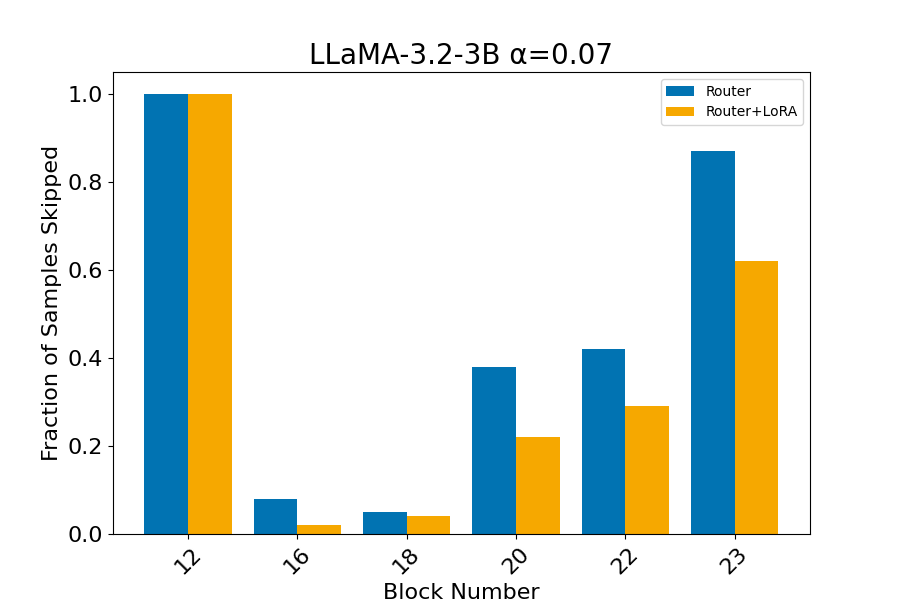}
  \end{minipage}%
  \hspace{0.05cm}  % Horizontal space between the top images
  \begin{minipage}{0.48\linewidth}
    \centering
    \includegraphics[width=\linewidth]{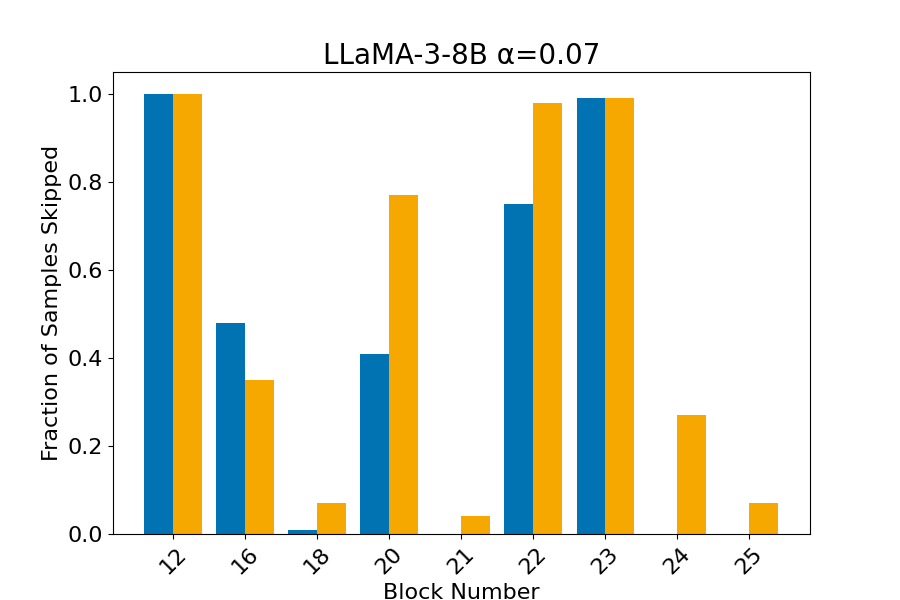}
  \end{minipage}
  \caption{Comparison of LLaMA-3.2-3B (left) and LLaMA-3-8B (right) at 10\% skipping rate on Natural Questions Answering Task. Layers with no skipping, indicated by a 0\% skipping rate, are not represented in the plot.}
  \label{fig:skipping_nq_10}
\end{figure*}

\subsection{Prompt Details}
\label{app:prompt_details}
\noindent The prompt structures used for both training and inference are as follows:

\begin{itemize}
    \item For the machine translation task (English-to-German or English-to-Chinese), the following general prompt structure is used to train the routers and during final inference:

    \small
    \begin{verbatim}
### Instruction:
    Translate the following sentences from English 
to German.
    
### Input:
    {Text to be translated}

### Response:
    \end{verbatim}
    \normalsize

    \item For the summarization task (used in CNN/DailyMail dataset), the prompt structure used is:
    \small
    \begin{verbatim}
### Instruction:
    Summarize the news article in around 100-200
words.

### Input:
    {Article to be summarized}

### Response:

    \end{verbatim}
    \normalsize
    \item For the Question Answering task (used in SQuAD/NQ dataset), the following prompt structure is utilized:

    \small
    \begin{verbatim}
### Instruction:
    Answer the question based on the given passage.

### Passage:
    {context}
    
### Question:
    {Question to be answered}
### Response:
    \end{verbatim}
    \normalsize

\end{itemize}

\noindent During the training of the LoRA module, task-aware training is applied. The expected translation or summary is appended after the \#\#\# Response section, making the model predict the response tokens following the "Response:\textbackslash n".

\subsection{Layer-wise Skipping Statistics} 
\label{app:skipping_stats}

% The router is trained in a task-aware manner, ensuring that skipping decisions align with task complexity. Our results demonstrate that these trained routers generalize effectively (see Appendix: \ref{app:generalizability and Reusability of Routers}. 
%Specifically, Table~\ref{tab:nq_mt_side_by_side_cross} evaluates the reusability and generalizability of routers across different datasets for the same underlying task. For instance, it shows Exact Match (EM) on Natural Questions (NQ) where routers trained on SQuAD (\textit{Cross-dataset}) are compared to those trained directly on NQ (\textit{Original}). Similarly, for English$\rightarrow$Chinese translation, routers trained on English$\rightarrow$German (\textit{Cross-dataset}) are evaluated against those trained on the target WMT direction (\textit{Original}). 
% The findings in Table~\ref{tab:nq_mt_side_by_side_cross} indicate strong cross-dataset generalization capabilities.
To illustrate broader layer-wise behaviors, Figures~\ref{fig:skipping_eng_de_15} through \ref{fig:skipping_nq_10} show block-wise skip rates observed when the models were configured for an average skip rate of approximately 15\%. These figures highlight how individual layers exhibit distinct skipping behaviors across various tasks, including translation (English-German, English-Chinese), summarization (CNN/DM), and question answering (SQuAD, Natural Questions). It also reveals that tasks with similar processing requirements (e.g., different translation language pairs) exhibit comparable skipping patterns.
Providing a more granular view for specific models, Table~\ref{tab:german_skip_8_formatted} reports the fraction of sequences that skip each block under LoRA adaptation for the LLaMA-3-8B and LLaMA-3.2-3B models. These fractions are presented as decimals, so a value of 0.10 signifies that 10\% of sequences skip the corresponding block.

\begin{table}
\centering
\tiny
\setlength{\tabcolsep}{3pt}
\begin{tabular}{c|c|c|c|c|c|c}
\toprule
\textbf{Skip (\%)} & \textbf{Model Type} & \textbf{Config} &  \textbf{BLEU-1}& \textbf{BLEU-2}&\textbf{BLEU} & \textbf{COMET} \\
\midrule
\multicolumn{7}{c}{\textbf{LLaMA-3.2-3B}} \\
\midrule
\multirow{2}{*}{\textbf{0}}  & \multirow{2}{*}{\textbf{Original Model}} & \textbf{Base + LoRA}   &        \textbf{0.376}& \textbf{0.174}&\textbf{0.160}& \textbf{89.72} \\
 &  & \textbf{Base}         &        0.312& 0.137&0.125& 81.66      \\
\midrule
\multirow{8}{*}{\textbf{15}} & \multirow{2}{*}{\textbf{Skip Decode}}  & \textbf{Router + LoRA} &        0.207& 0.069&0.105& 44.58      \\
 &  & \textbf{Router}       &        0.099& 0.039&0.055& 32.74      \\
 & \multirow{2}{*}{\textbf{Random Skip}}  & \textbf{Router + LoRA} &        0.190& 0.048&0.079& 47.26      \\
 &  & \textbf{Router}       &        0.121& 0.028&0.048& 36.30      \\
 & \multirow{2}{*}{\textbf{Unified Skip}} & \textbf{Router + LoRA} &        0.216& 0.059&0.095& 44.72      \\
 &  & \textbf{Router}       &        0.164& 0.040&0.067& 39.81      \\
 & \multirow{2}{*}{\textbf{FiRST (Ours)}}  & \textbf{Router + LoRA} &        \textbf{0.247}& \textbf{0.071}&\textbf{0.113}& \textbf{60.29} \\
 &  & \textbf{Router}       &        0.165& 0.041&0.069& 43.04      \\
\midrule
\multirow{8}{*}{\textbf{25}} & \multirow{2}{*}{\textbf{Skip Decode}}  & \textbf{Router + LoRA} &        0.168& \textbf{0.047}&\textbf{0.077}& 32.33      \\
 &  & \textbf{Router}       &        0.091& 0.035&0.050& 27.64      \\
 & \multirow{2}{*}{\textbf{Random Skip}}  & \textbf{Router + LoRA} &        0.090& 0.011&0.023& 30.97      \\
 &  & \textbf{Router}       &        0.053& 0.007&0.015& 27.22      \\
 & \multirow{2}{*}{\textbf{Unified Skip}} & \textbf{Router + LoRA} &        0.151& 0.027&0.050& 39.81      \\
 &  & \textbf{Router}       &        0.102& 0.016&0.032& 30.86      \\
 & \multirow{2}{*}{\textbf{FiRST (Ours)}}  & \textbf{Router + LoRA} &        \textbf{0.189}& 0.041&0.072& \textbf{45.38} \\
 &  & \textbf{Router}       &        0.095& 0.014&0.029& 29.08      \\
\midrule
\multirow{8}{*}{\textbf{35}} & \multirow{2}{*}{\textbf{Skip Decode}}  & \textbf{Router + LoRA} &        0.058& 0.011&0.018& 23.30      \\
 &  & \textbf{Router}       &        0.017& 0.003&0.005& 19.34      \\
 & \multirow{2}{*}{\textbf{Random Skip}}  & \textbf{Router + LoRA} &        0.047& 0.001&0.007& 25.08      \\
 &  & \textbf{Router}       &        0.020& 0.001&0.003& 21.18      \\
 & \multirow{2}{*}{\textbf{Unified Skip}} & \textbf{Router + LoRA} &        0.012& 0.000&0.002& 19.65      \\
 &  & \textbf{Router}       &        0.009& 0.000&0.001& 20.36      \\
 & \multirow{2}{*}{\textbf{FiRST (Ours)}}  & \textbf{Router + LoRA} &      \textbf{ 0.095}& \textbf{0.013}&\textbf{0.027}& \textbf{27.45} \\
 &  & \textbf{Router}       &        0.057& 0.006&0.014& 25.03      \\
\midrule
  \multicolumn{7}{c}{\textbf{LLaMA-3-8B}} \\
\midrule
\multirow{2}{*}{\textbf{0}}  & \multirow{2}{*}{\textbf{Original Model}} & \textbf{Base + LoRA} &  \textbf{0.418}& \textbf{0.217}&\textbf{0.199}& \textbf{93.00} \\
 &  & \textbf{Base}         &  0.372& 0.186&0.169& 87.13      \\
\midrule
\multirow{8}{*}{\textbf{15}} & \multirow{2}{*}{\textbf{Skip Decode}}  & \textbf{Router + LoRA} &  0.230& 0.105&0.094& 55.62      \\
 &  & \textbf{Router}       &  0.040& 0.012&0.019& 23.33      \\
 & \multirow{2}{*}{\textbf{Random Skip}}  & \textbf{Router + LoRA} &  0.304& 0.110&0.097& 66.25      \\
 &  & \textbf{Router}       &  0.265& 0.088&0.132& 60.27      \\
 & \multirow{2}{*}{\textbf{Unified Skip}} & \textbf{Router + LoRA} &  0.289& 0.106&0.153& 59.34      \\
 &  & \textbf{Router}       &  0.232& 0.079&0.117& 59.26      \\
 & \multirow{2}{*}{\textbf{FiRST (Ours)}}  & \textbf{Router + LoRA} &  \textbf{0.380}& \textbf{0.179}&\textbf{0.161}& \textbf{82.14} \\
 &  & \textbf{Router}       &  0.288& 0.118&0.108& 67.74      \\
\midrule
\multirow{8}{*}{\textbf{25}} & \multirow{2}{*}{\textbf{Skip Decode}}  & \textbf{Router + LoRA} &  0.118& \textbf{0.052}&0.070& 31.47      \\
 &  & \textbf{Router}       &  0.032& 0.009&0.015& 21.55      \\
 & \multirow{2}{*}{\textbf{Random Skip}}  & \textbf{Router + LoRA} &  0.060& 0.009&0.018& 29.71      \\
 &  & \textbf{Router}       &  0.037& 0.005&0.011& 29.95      \\
 & \multirow{2}{*}{\textbf{Unified Skip}} & \textbf{Router + LoRA} &  0.157& 0.034&0.060& 31.69      \\
 &  & \textbf{Router}       &  0.126& 0.027&0.048& 32.15      \\
 & \multirow{2}{*}{\textbf{FiRST (Ours)}}  & \textbf{Router + LoRA} & \textbf{ 0.178}& 0.041&\textbf{0.071}& \textbf{34.95} \\
 &  & \textbf{Router}       &  0.097& 0.014&0.029& 26.01      \\
\midrule
\multirow{8}{*}{\textbf{35}} & \multirow{2}{*}{\textbf{Skip Decode}}  & \textbf{Router + LoRA} &  0.049& \textbf{0.020}&\textbf{0.028}& 23.85      \\
 &  & \textbf{Router}       &  0.030& 0.008&0.013& 20.03      \\
 & \multirow{2}{*}{\textbf{Random Skip}}  & \textbf{Router + LoRA} &  0.018& 0.001&0.004& \textbf{25.56} \\
 &  & \textbf{Router}       &  0.014& 0.001&0.002& 25.34      \\
 & \multirow{2}{*}{\textbf{Unified Skip}} & \textbf{Router + LoRA} &  \textbf{0.064}& 0.008&0.017& 22.05      \\
 &  & \textbf{Router}       &  0.039& 0.005&0.011& 22.88      \\
 & \multirow{2}{*}{\textbf{FiRST (Ours)}}  & \textbf{Router + LoRA} &  \textbf{0.064}& 0.004&0.012& 19.96      \\
 &  & \textbf{Router}       &  0.037& 0.001&0.005& 21.41      \\
\bottomrule
\end{tabular}
\caption{\small Machine Translation Results for English to German on LLaMA-3.2-3B and LLaMA-3-8B: BLEU and COMET scores for various skipping strategies.}
\label{tab:translation_results_eng_de}
\end{table}

\begin{table}
\centering
\tiny
\setlength{\tabcolsep}{3pt}
\begin{tabular}{c|c|c|c|c|c|c}
\toprule
\textbf{Skip (\%)} & \textbf{Model Type} & \textbf{Config} & \textbf{BLEU-1} & \textbf{BLEU-2} & \textbf{BLEU} & \textbf{COMET} \\
\midrule
\multicolumn{7}{c}{\textbf{LLaMA-3.2-3B}} \\
\midrule
\multirow{2}{*}{\textbf{0}}  & \multirow{2}{*}{\textbf{Original Model}} & \textbf{Base + LoRA} &      \textbf{0.518}&      \textbf{0.300}&      \textbf{0.278}& \textbf{79.13} \\
 &  & \textbf{Base}            &      0.321&      0.179&      0.166& 61.84         \\
\midrule
\multirow{8}{*}{\textbf{15}} & \multirow{2}{*}{\textbf{Skip Decode}}  & \textbf{Router + LoRA} &      0.381&      0.217&      0.200& 46.70         \\
 &  & \textbf{Router}          &      0.096&      0.048&      0.062& 34.14         \\
 & \multirow{2}{*}{\textbf{Random Skip}}  & \textbf{Router + LoRA} &      0.387&      0.174&      0.154& 57.79         \\
 &  & \textbf{Router}          &      0.136&      0.056&      0.077& 35.86         \\
 & \multirow{2}{*}{\textbf{Unified Skip}} & \textbf{Router + LoRA} &      0.370&      0.173&      0.157& 57.10         \\
 &  & \textbf{Router}          &      0.224&      0.094&      0.087& 45.16         \\
 & \multirow{2}{*}{\textbf{FiRST (Ours)}}  & \textbf{Router + LoRA} &      \textbf{0.457}&      \textbf{0.237}&      \textbf{0.218}& \textbf{67.45} \\
 &  & \textbf{Router}          &      0.227&      0.109&      0.100& 54.55         \\
\midrule
\multirow{8}{*}{\textbf{25}} & \multirow{2}{*}{\textbf{Skip Decode}}  & \textbf{Router + LoRA} &      0.278&      \textbf{0.157}&      \textbf{0.147}& 42.01         \\
 &  & \textbf{Router}          &      0.070&      0.034&      0.045& 29.51         \\
 & \multirow{2}{*}{\textbf{Random Skip}}  & \textbf{Router + LoRA} &      0.251&      0.099&      0.137& \textbf{45.53} \\
 &  & \textbf{Router}          &      0.101&      0.037&      0.053& 31.50         \\
 & \multirow{2}{*}{\textbf{Unified Skip}} & \textbf{Router + LoRA} &      0.306&      0.123&      0.108& 42.30         \\
 &  & \textbf{Router}          &      0.152&      0.055&      0.079& 31.63         \\
 & \multirow{2}{*}{\textbf{FiRST (Ours)}}  & \textbf{Router + LoRA} &      \textbf{0.329}&      0.137&      0.126& 41.66         \\
 &  & \textbf{Router}          &      0.105&      0.035&      0.053& 27.83         \\
\midrule
\multirow{8}{*}{\textbf{35}} & \multirow{2}{*}{\textbf{Skip Decode}}  & \textbf{Router + LoRA} &      0.028&      0.011&      0.016& 22.30         \\
 &  & \textbf{Router}          &      0.024&      0.010&      0.014& 18.32         \\
 & \multirow{2}{*}{\textbf{Random Skip}}  & \textbf{Router + LoRA} &      0.042&      0.015&      0.022& \textbf{32.81} \\
 &  & \textbf{Router}          &      0.021&      0.008&      0.011& 23.63         \\
 & \multirow{2}{*}{\textbf{Unified Skip}} & \textbf{Router + LoRA} &      0.033&      0.005&      0.010& 19.12         \\
 &  & \textbf{Router}          &      0.022&      0.004&      0.007& 19.82         \\
 & \multirow{2}{*}{\textbf{FiRST (Ours)}}  & \textbf{Router + LoRA} &      \textbf{0.100}&      \textbf{0.070}&      \textbf{0.075}& 28.10         \\
 &  & \textbf{Router}          &      0.065&      0.018&      0.029& 23.13         \\
\midrule
\multicolumn{7}{c}{\textbf{LLaMA-3-8B}} \\
\midrule
\multirow{2}{*}{\textbf{0}}  & \multirow{2}{*}{\textbf{Original Model}} & \textbf{Base + LoRA}      &      \textbf{0.569}&      \textbf{0.356}& \textbf{0.333}& \textbf{82.66} \\
 &  & \textbf{Base}            &      0.380&      0.225& 0.208& 68.95         \\
\midrule
\multirow{8}{*}{\textbf{15}} & \multirow{2}{*}{\textbf{Skip Decode}}  & \textbf{Router + LoRA} &      0.287&      0.158& 0.149& 55.98         \\
 &  & \textbf{Router}          &      0.047&      0.023& 0.030& 21.75         \\
 & \multirow{2}{*}{\textbf{Random Skip}}  & \textbf{Router + LoRA} &      0.479&      0.258& 0.237& 67.32         \\
 &  & \textbf{Router}          &      0.366&      0.187& 0.168& 59.89         \\
 & \multirow{2}{*}{\textbf{Unified Skip}} & \textbf{Router + LoRA} &      0.466&      0.250& 0.230& \textbf{69.58} \\
 &  & \textbf{Router}          &      0.273&      0.134& 0.122& 54.57         \\
 & \multirow{2}{*}{\textbf{FiRST (Ours)}}  & \textbf{Router + LoRA} &      \textbf{0.484}&      \textbf{0.266}& \textbf{0.247}& 68.63         \\
 &  & \textbf{Router}          &      0.176&      0.087& 0.080& 42.76         \\
\midrule
\multirow{8}{*}{\textbf{25}} & \multirow{2}{*}{\textbf{Skip Decode}}  & \textbf{Router + LoRA} &      0.173&      0.094& 0.088& 33.85         \\
 &  & \textbf{Router}          &      0.038&      0.018& 0.024& 20.93         \\
 & \multirow{2}{*}{\textbf{Random Skip}}  & \textbf{Router + LoRA} &      0.117&      0.047& 0.065& 27.73         \\
 &  & \textbf{Router}          &      0.074&      0.028& 0.040& 35.16         \\
 & \multirow{2}{*}{\textbf{Unified Skip}} & \textbf{Router + LoRA} &      0.349&      \textbf{0.158}& \textbf{0.142}& 50.59         \\
 &  & \textbf{Router}          &      0.177&      0.074& 0.068& 38.74         \\
 & \multirow{2}{*}{\textbf{FiRST (Ours)}}  & \textbf{Router + LoRA} &      \textbf{0.358}&	\textbf{0.157}& 0.119& \textbf{56.92} \\
 &  & \textbf{Router}          &      0.110&      0.032& 0.051& 25.45         \\
\midrule
\multirow{8}{*}{\textbf{35}} & \multirow{2}{*}{\textbf{Skip Decode}}  & \textbf{Router + LoRA} &      0.112&      \textbf{0.057}& 0.054& 25.23         \\
 &  & \textbf{Router}          &      0.036&      0.017& 0.023& 22.84         \\
 & \multirow{2}{*}{\textbf{Random Skip}}  & \textbf{Router + LoRA} &      0.074&      0.022& 0.034& 27.35         \\
 &  & \textbf{Router}          &      0.045&      0.011& 0.018& \textbf{29.46} \\
 & \multirow{2}{*}{\textbf{Unified Skip}} & \textbf{Router + LoRA} &      0.075&      0.021& 0.034& 20.25         \\
 &  & \textbf{Router}          &      0.039&      0.011& 0.017& 21.24         \\
 & \multirow{2}{*}{\textbf{FiRST (Ours)}}  & \textbf{Router + LoRA} &      \textbf{0.157}&      0.040& \textbf{0.066}& 26.80         \\
 &  & \textbf{Router}          &      0.061&      0.015& 0.025& 22.89         \\
\bottomrule
\end{tabular}
\caption{\small Machine Translation Results for English to Chinese on LLaMA-3.2-3B and LLaMA-3-8B: BLEU and COMET scores for various skipping strategies.}
\label{tab:translation_results_eng_zh}
\end{table}

\begin{table}
\centering
\tiny
\setlength{\tabcolsep}{3pt}

\begin{tabular}{@{} c  cc |  cc | cc @{}}
\hline\\[-1.2ex]
\textbf{Layer $\downarrow$} & \multicolumn{2}{c}{\textbf{$\alpha=0.005$}} & \multicolumn{2}{c}{\textbf{$\alpha=0.01$}} & \multicolumn{2}{c}{\textbf{$\alpha=0.025$}} \\
\cmidrule(lr){2-3} \cmidrule(lr){4-5} \cmidrule(lr){6-7}
 & \textbf{R} & \textbf{R+L} & \textbf{R} & \textbf{R+L} & \textbf{R} & \textbf{R+L} \\
\hline\\[-1.2ex]
\textbf{0}  & 0.00 & 0.00 & 0.00 & 0.00 & 0.00 & 0.00 \\
\textbf{1}  & 0.00 & 0.00 & 0.00 & 0.00 & 0.00 & 0.00 \\
\textbf{2}  & 0.00 & 0.00 & 0.00 & 0.00 & 0.00 & 0.00 \\
\textbf{3}  & 0.00 & 0.00 & 0.00 & 0.00 & 0.00 & 0.00 \\
\textbf{4}  & 0.00 & 0.00 & 0.00 & 0.00 & 0.00 & 0.00 \\
\textbf{5}  & 0.00 & 0.00 & 0.00 & 0.00 & 0.00 & 0.00 \\
\textbf{6}  & 0.00 & 0.00 & 0.00 & 0.00 & 0.00 & 0.00 \\
\textbf{7}  & 100.00 & 100.00 & 100.00 & 100.00 & 100.00 & 100.00 \\
\textbf{8}  & 100.00 & 100.00 & 100.00 & 100.00 & 100.00 & 100.00 \\
\textbf{9}  & 100.00 & 100.00 & 0.00 & 0.00 & 0.10 & 1.32 \\
\textbf{10} & 0.00 & 0.00 & 0.00 & 0.00 & 89.25 & 72.67 \\
\textbf{11} & 0.00 & 0.00 & 0.00 & 0.00 & 0.00 & 0.00 \\
\textbf{12} & 0.00 & 0.00 & 100.00 & 100.00 & 100.00 & 100.00 \\
\textbf{13} & 0.00 & 0.00 & 0.00 & 0.00 & 0.00 & 0.00 \\
\textbf{14} & 0.00 & 0.00 & 0.00 & 0.59 & 0.00 & 0.25 \\
\textbf{15} & 0.00 & 0.00 & 100.00 & 99.95 & 100.00 & 99.36 \\
\textbf{16} & 0.00 & 0.00 & 0.10 & 2.45 & 100.00 & 99.95 \\
\textbf{17} & 0.00 & 0.00 & 0.00 & 0.00 & 0.00 & 0.00 \\
\textbf{18} & 97.79 & 32.24 & 100.00 & 100.00 & 100.00 & 100.00 \\
\textbf{19} & 0.00 & 0.00 & 99.85 & 91.17 & 100.00 & 99.61 \\
\textbf{20} & 0.00 & 0.00 & 100.00 & 100.00 & 100.00 & 99.46 \\
\textbf{21} & 99.85 & 98.72 & 100.00 & 100.00 & 100.00 & 100.00 \\
\textbf{22} & 0.00 & 0.00 & 0.00 & 0.00 & 0.00 & 0.00 \\
\textbf{23} & 0.00 & 0.00 & 24.14 & 0.79 & 99.75 & 91.66 \\
\textbf{24} & 0.00 & 0.00 & 0.00 & 0.00 & 0.00 & 0.00 \\
\textbf{25} & 0.00 & 0.00 & 0.00 & 0.00 & 0.00 & 0.00 \\
\textbf{26} & 0.00 & 0.00 & 57.31 & 2.45 & 100.00 & 86.02 \\
\textbf{27} & 0.00 & 0.00 & 0.00 & 0.00 & 0.00 & 0.00 \\
\textbf{28} & 0.00 & 0.00 & 0.00 & 0.00 & 0.00 & 0.05 \\
\textbf{29} & 0.00 & 0.00 & 0.00 & 0.00 & 0.00 & 0.00 \\
\textbf{30} & 0.00 & 0.00 & 0.00 & 0.00 & 0.00 & 0.00 \\
\textbf{31} & 0.00 & 0.00 & 0.00 & 0.00 & 0.00 & 0.00 \\
\textbf{Avg} & 15.55 & 13.47 & 27.54 & 24.92 & 37.16 & 35.95 \\
\hline
\end{tabular}

\caption{\small Variation in skipping percentage (15–35\%) with the non-skip penalization loss coefficient $\alpha$ for LLaMA-3-8B on English–German translation. As $\alpha$ increases, average skipping rises across both Router-only (R) and Router+LoRA (R+L) models.}
\label{tab:german_skip_8_formatted}
\end{table}

% For LLaMA-3.2-3B, the latency improvement is $\sim 10\%$, with quality scores being significantly higher than other layer skipping strategies (Table \ref{tab:tpot_total_3}). It is within $65 - 85\%$ of BLEU-1 and COMET scores (Tables \ref{tab:translation_results_3_eng_de} and \ref{tab:translation_results_3_eng_zh}) for EN-DE and EN-ZH of the gold (LoRA fine-tuned base model). 

% \noindent \textbf{CNN/DailyMail Dataset:} For LLaMA-3-8B, at roughly 15\% skipping level, our method \textbf{outperforms} the base model + LoRA (Table: \ref{tab:summ_combined_results}) while obtaining a 12\% improvement in TPOT (Table \ref{tab:tpot_total_8}).   
% For LLaMA-3.2-3B, at $15\%$, the quality is comparable ($\sim 98\%$) to gold and other baselines with $12\%$ improvement in TPOT. At $24\%$, \textsc{FiRST} is significantly better than other layer skipping strategies, while achieving $>20\%$ improvement in latency. 

\begin{table*}
\centering
\tiny
\begin{tabular}{ c | c | c | c | c | c }
\toprule
\textbf{Skip (\%)} & \multicolumn{2}{c}{\textbf{Model Type}} & \textbf{BERT} & \textbf{R-1} & \textbf{R-L} \\
\toprule
\multirow{2}{*}{\textbf{0}} & \multirow{2}{*}{\textbf{Original Model}} & \textbf{wLoRA}& \textbf{84.87} & \textbf{28.46} & \textbf{16.99} \\
 &  & \textbf{Base} & 82.29 & 23.49 & 14.66 \\
\midrule
\multirow{8}{*}{\textbf{15}} & \multirow{2}{*}{\textbf{Skip Decode}} & \textbf{R + LoRA}& 84.74 & 22.04 & 17.54 \\
 &  & \textbf{Router } & 82.53 & 13.68 & 9.30 \\
 & \multirow{2}{*}{\textbf{Random Skip}} & \textbf{R + LoRA}& 83.70 & 24.60 & 15.01 \\
 &  & \textbf{Router } & 81.10 & 19.64 & 13.07 \\
 & \multirow{2}{*}{\textbf{Unified Skip}} & \textbf{R + LoRA}& 84.25 & 24.35 & 14.3 \\
 &  & \textbf{Router } & 80.3 & 16.61 & 10.95 \\
 & \multirow{2}{*}{\textbf{FiRST (Ours)}} & \textbf{R + LoRA}& \textbf{85.14} & \textbf{31.8} & \textbf{20.13} \\
 &  & \textbf{Router } & 81.25 & 20.2 & 13.01 \\
 \midrule
\multirow{8}{*}{\textbf{20}} & \multirow{2}{*}{\textbf{Skip Decode}} & \textbf{R + LoRA} & 82.57 & 20.41 & 14.87 \\
 &  & \textbf{Router } & 81.62 & 13.48 & 9.19 \\
 & \multirow{2}{*}{\textbf{Random Skip}} & \textbf{R + LoRA} & 81.39 & 21.57 & 13.83 \\
 &  & \textbf{Router } & 79.23 & 15.51 & 10.93 \\
 & \multirow{2}{*}{\textbf{Unified Skip}} & \textbf{R + LoRA} & \textbf{82.93} & 22.3 & 13.37 \\
 &  & \textbf{Router } & 80.32 & 16.51 & 11.15 \\
 & \multirow{2}{*}{\textbf{FiRST (Ours)}} & \textbf{R + LoRA} & 82.8 & \textbf{27.65} & \textbf{17.84} \\
 &  & \textbf{Router } & 79.32 & 16.28 & 10.85 \\
 \midrule
\multirow{8}{*}{\textbf{27}} & \multirow{2}{*}{\textbf{Skip Decode}} & \textbf{R + LoRA}& 79.92 & 10.67 & 10.32 \\
 &  & \textbf{Router } & 77.27 & 9.59 & 7.00 \\
 & \multirow{2}{*}{\textbf{Random Skip}} & \textbf{R + LoRA}& 76.40 & 11.45 & 7.89 \\
 &  & \textbf{Router } & 77.45 & 12.56 & 9.08 \\
 & \multirow{2}{*}{\textbf{Unified Skip}} & \textbf{R + LoRA}& \textbf{80.28} & \textbf{15.94} & 9.89 \\
 &  & \textbf{Router } & 77.43 & 10.97 & 7.68 \\
 & \multirow{2}{*}{\textbf{FiRST (Ours)}} & \textbf{R + LoRA}& 77.5 & 14.65 & \textbf{10.45} \\
 &  & \textbf{Router } & 75.6 & 9.39 & 6.92 \\
\bottomrule
\end{tabular}
\hspace{1cm}
\begin{tabular}{ c | c | c | c | c | c }
\toprule
\textbf{Skip (\%)} & \multicolumn{2}{c}{\textbf{Model Type}} & \textbf{BERT} & \textbf{R-1} & \textbf{R-L} \\
\toprule
\multirow{2}{*}{\textbf{0}} & \multirow{2}{*}{\textbf{Original Model}} & \textbf{wLoRA} & \textbf{84.89} & \textbf{28.37} & \textbf{17.02} \\
 &  & \textbf{Base} & 71.85 & 19.34 & 12.00 \\
 \midrule
\multirow{8}{*}{\textbf{15}} & \multirow{2}{*}{\textbf{Skip Decode}} & \textbf{R + LoRA} & \textbf{83.20} & 21.71 & 13.74 \\
 &  & \textbf{Router } & 80.97 & 9.74 & 6.87 \\
 & \multirow{2}{*}{\textbf{Random Skip}} & \textbf{R + LoRA} & 79.52 & 20.18 & 12.10 \\
 &  & \textbf{Router } & 68.20 & 10.10 & 7.10 \\
 & \multirow{2}{*}{\textbf{Unified Skip}} & \textbf{R + LoRA} & 81.53 & 18.89 & 11.72 \\
 &  & \textbf{Router } & 70.01 & 12.49 & 8.68 \\
 & \multirow{2}{*}{\textbf{FiRST (Ours)}} & \textbf{R + LoRA} & 83.17 & \textbf{26.47} & \textbf{16.79} \\
 &  & \textbf{Router } & 70.98 & 16.47 & 10.51 \\
 \midrule
\multirow{8}{*}{\textbf{24}} & 
\multirow{2}{*}{\textbf{Skip Decode}} & \textbf{R + LoRA} & 78.55 & 15.83 & 6.74 \\
 &  & \textbf{Router } & 76.91 & 13.29 & 8.86 \\
 & \multirow{2}{*}{\textbf{Random Skip}} & \textbf{R + LoRA} & 80.00 & 16.33 & 10.07 \\
 &  & \textbf{Router } & 67.88 & 8.49 & 5.96 \\
 & \multirow{2}{*}{\textbf{Unified Skip}} & \textbf{R + LoRA} & 79.31 & 15.88 & 10.69 \\
 &  & \textbf{Router } & 68.86 & 9.17 & 6.97 \\
 & \multirow{2}{*}{\textbf{FiRST (Ours)}} & \textbf{R + LoRA} & \textbf{80.25} & \textbf{21.28} & \textbf{13.89} \\
 &  & \textbf{Router } & 69.17 & 12.36 & 8.43 \\
 \midrule
\multirow{8}{*}{\textbf{28}} & 
\multirow{2}{*}{\textbf{Skip Decode}} & \textbf{R + LoRA} & 70.69& 8.76& 6.74\\
 &  & \textbf{Router } & 40.99& 2.05& 1.23\\
 & \multirow{2}{*}{\textbf{Random Skip}} & \textbf{R + LoRA} & \textbf{79.45}& 14.69& 9.23\\
 &  & \textbf{Router } & 67.48& 8.14& 5.64\\
 & \multirow{2}{*}{\textbf{Unified Skip}} & \textbf{R + LoRA} & 78.57& 11.74& 7.47\\
 &  & \textbf{Router } & 68.23& 8.12& 5.66\\
 & \multirow{2}{*}{\textbf{FiRST (Ours)}} & \textbf{R + LoRA} & 77.48& \textbf{15.98}& \textbf{11.14}\\
 &  & \textbf{Router } & 67.14& 8.09& 6.00\\
\bottomrule
\end{tabular}
\caption{Quality Analysis on Summarization (CNN/DM dataset) on LLaMA-3-8B (left) and LLaMA-3.2-3B (right): BERT F1, Rouge-1 and Rouge-L scores are reported for varying skipping levels. Note that R + LoRA corresponds to Router Augmentation followed by LoRA fine-tuning (in the proposed FiRST framework) and wLoRA stands for Base Model with LoRA fine-tuning. 
FiRST with fine-tuning, improves upon Unified Skipping for all skipping levels on both Rouge-1 and Rouge-L and is competitive on BERT F1.}
\label{tab:cnn_dm_extended}
\end{table*}

% \begin{table*}
% \centering
% \scriptsize
% \begin{tabular}{ c |c |c| c}
% \toprule
% \textbf{Model Type} & \textbf{$\sim$ Skipping (\%)} &\textbf{LLaMA-3.2-3B} & \textbf{LLaMA-3-8B}\\
% \textbf{} & \textbf{} &\textbf{TPOT}  &\textbf{TPOT} \\
% \toprule
% \textbf{Base + LoRA} & 0&1x &1x \\
% \midrule
% \textbf{R + LoRA} & 15&0.91x &0.88x\\
% \midrule
% \textbf{R + LoRA} & 25&0.75x &0.83x\\
% \midrule
% \textbf{R + LoRA} & 35&0.74x &0.68x\\
% \bottomrule
% \end{tabular}

% \caption{\small{TPOT variations for LLaMA-3.2-3B (left) and LLaMA-3-8B (right) on the English-to-Chinese WMT task (FiRST): The reported values are relative to the LoRA fine-tuned base model. Fine-tuning improves TPOT and quality significantly.}}
% \label{tab:tpot_total_eng_zh}
% \end{table*}

% \noindent \textbf{SQuAD Dataset:}
% For the LLaMA-3-8B model, \textsc{FiRST} is $> 95\%$ in output quality of gold (base + LoRA) (Table \ref{tab:extended_squad}), with overall latency gains of 6-16\% (Table \ref{tab:tpot_squad_nq}). It is \textbf{significantly} better in quality than all other baselines across all metrics for different levels of skipping. 

% For LLaMA-3.2-3B, again \textsc{FiRST} is $> 95\%$ in output quality of gold (base + LoRA) for 10\% skipping (Table \ref{tab:extended_squad}) with gains in latency of 6-16\% overall (Table \ref{tab:tpot_squad_nq}) over the LoRA fine-tuned base model. Moreover, it is better than all other layer skipping strategies across all metrics. 

\begin{table*}
\centering
\scriptsize
\begin{tabular}{ c | c | c | c | c }
\toprule
\textbf{Skip (\%)} & \multicolumn{2}{c|}{\textbf{Model Type}} & \textbf{EM} & \textbf{F1} \\
\toprule
\multirow{2}{*}{\textbf{0}} & \multirow{2}{*}{\textbf{Original Model}} & \textbf{wLoRA} & \textbf{73.93} & \textbf{85.99} \\
 &  & \textbf{Base} & 19.46 & 36.73 \\
 \midrule
\multirow{8}{*}{\textbf{10}} & \multirow{2}{*}{\textbf{Skip Decode}} & \textbf{R + LoRA}& 60.14 & 65.33 \\
 &  & \textbf{Router } & 16.38 & 31.48 \\
 & \multirow{2}{*}{\textbf{Random Skip}} & \textbf{R + LoRA}& 65.73 & 80.08 \\
 &  & \textbf{Router } & 18.25 & 33.75 \\
 & \multirow{2}{*}{\textbf{Unified Skip}} & \textbf{R + LoRA}& 55.54 & 74.58 \\
 &  & \textbf{Router } & 17.39 & 32.91 \\
 & \multirow{2}{*}{\textbf{FiRST (Ours)}} & \textbf{R + LoRA}& \textbf{70.85} & \textbf{83.61} \\
 &  & \textbf{Router } & 14.58 & 31.52 \\
 \midrule
\multirow{8}{*}{\textbf{20}} & \multirow{2}{*}{\textbf{Skip Decode}} & \textbf{R + LoRA}& 45.00 & 55.10 \\
 &  & \textbf{Router } & 10.68 & 26.69 \\
 & \multirow{2}{*}{\textbf{Random Skip}} & \textbf{R + LoRA}& 47.79 & 66.37 \\
 &  & \textbf{Router } & 6.71 & 22.46 \\
 & \multirow{2}{*}{\textbf{Unified Skip}} & \textbf{R + LoRA}& 52.87 & 69.28 \\
 &  & \textbf{Router } & 18.18 & 32.51 \\
 & \multirow{2}{*}{\textbf{FiRST (Ours)}} & \textbf{R + LoRA}& \textbf{60.60} & \textbf{75.49} \\
 &  & \textbf{Router } & 13.21 & 27.48 \\
 \midrule
\multirow{8}{*}{\textbf{30}} & \multirow{2}{*}{\textbf{Skip Decode}} & \textbf{R + LoRA} & 30.77 & 48.38 \\

 &  & \textbf{Router } & 10.67 & 28.52 \\
 & \multirow{2}{*}{\textbf{Random Skip}} & \textbf{R + LoRA} & 25.45 & 42.68 \\
 &  & \textbf{Router } & 3.55 & 15.49 \\
 & \multirow{2}{*}{\textbf{Unified Skip}} & \textbf{R + LoRA} & 25.61 & 38.55 \\
 &  & \textbf{Router } & 15.19 & 28.11 \\
 & \multirow{2}{*}{\textbf{FiRST (Ours)}} & \textbf{R + LoRA} & \textbf{38.20} & \textbf{52.68} \\
&  & \textbf{Router } & 3.64 & 13.30 \\
 \bottomrule
\end{tabular}
\hspace{1cm}
\begin{tabular}{ c | c | c | c | c }
\toprule
\textbf{Skip (\%)} & \multicolumn{2}{c|}{\textbf{Model Type}} & \textbf{EM} & \textbf{F1} \\
\toprule
\multirow{2}{*}{\textbf{0}} & \multirow{2}{*}{\textbf{Original Model}} & \textbf{wLoRA} & \textbf{73.07} & \textbf{84.17} \\
 &  & \textbf{Base} & 18.92 & 37.74 \\
 \midrule
\multirow{8}{*}{\textbf{10}} & \multirow{2}{*}{\textbf{Skip Decode}} & \textbf{R + LoRA} & 60.79 & 75.00 \\
 &  & \textbf{Router } & 20.00 & 31.55 \\
 & \multirow{2}{*}{\textbf{Random Skip}} & \textbf{R + LoRA} & 64.78 & 77.27 \\
 &  & \textbf{Router } & 13.76 & 28.59 \\
 & \multirow{2}{*}{\textbf{Unified Skip}} & \textbf{R + LoRA} & 65.03 & 77.53 \\
 &  & \textbf{Router } & 13.16 & 32.31 \\
 & \multirow{2}{*}{\textbf{FiRST}} & \textbf{R + LoRA} & \textbf{69.44} & \textbf{81.35} \\
 &  & \textbf{Router } & 12.79 & 28.37 \\
 \midrule
\multirow{8}{*}{\textbf{20}} & \multirow{2}{*}{\textbf{Skip Decode}} & \textbf{R + LoRA} & \textbf{40.12} & 40.00 \\
 &  & \textbf{Router } & 20.45 & 37.62 \\
 & \multirow{2}{*}{\textbf{Random Skip}} & \textbf{R + LoRA} & 11.32 & 38.34 \\
 &  & \textbf{Router } & 6.75 & 15.51 \\
 & \multirow{2}{*}{\textbf{Unified Skip}} & \textbf{R + LoRA} & 37.39 & 52.49 \\
 &  & \textbf{Router } & 7.81 & 18.20 \\
 & \multirow{2}{*}{\textbf{FiRST}} & \textbf{R + LoRA} & 39.70 & \textbf{54.59} \\
 &  & \textbf{Router } & 5.52 & 15.33 \\
 \midrule
 \multirow{8}{*}{\textbf{30}} & \multirow{2}{*}{\textbf{Skip Decode}} & \textbf{R + LoRA} & 20.68& 23.08\\
 &  & \textbf{Router } & 0.78& 3.83\\
 & \multirow{2}{*}{\textbf{Random Skip}} & \textbf{R + LoRA} & 0.30& 8.87\\
 &  & \textbf{Router } & 0.62& 5.88\\
 & \multirow{2}{*}{\textbf{Unified Skip}} & \textbf{R + LoRA} & 7.39& 13.72\\
 &  & \textbf{Router } & 0.37& 6.15\\
 & \multirow{2}{*}{\textbf{FiRST}} & \textbf{R + LoRA} & \textbf{33.99}& \textbf{50.37}\\
 &  & \textbf{Router } & 2.55& 10.26\\
\bottomrule
\end{tabular}
\caption{SQuAD performance on LLaMA-3-8B (left) and LLaMA-3.2-3B (right): EM (Exact Match) and F1 scores are reported for varying skipping levels. Note that R + LoRA corresponds to Router Augmentation followed by LoRA fine-tuning (in the proposed FiRST framework) and wLoRA stands for Base Model with LoRA fine-tuning. }
\label{tab:extended_squad}
\end{table*}

\begin{table*}
\renewcommand{\arraystretch}{1.05} % Slightly more vertical space
\centering
\scriptsize
\begin{tabular}{ c |c |c |c}
\toprule
\textbf{Model Type} & \textbf{$\sim$ Skipping (\%)} & \textbf{Eng$\rightarrow$De}&\textbf{Eng$\rightarrow$Zh}\\
\textbf{} & \textbf{} & \textbf{TPOT}   &\textbf{TPOT} \\
\hline
Base + LoRA & 0 & 1x   &1x \\
\hline
R + LoRA & 15 & 0.90x  &0.88x\\
\hline
R + LoRA & 25 & 0.82x  &0.83x\\
\hline
R + LoRA & 35 & 0.69x  &0.68x\\
\bottomrule
\end{tabular}
\hspace{1cm}
\begin{tabular}{ c  |c  |c }
\toprule
\textbf{Model Type} & \textbf{$\sim$Skipping (\%)} & \textbf{CNN/DM}  \\
\textbf{} & \textbf{} & \textbf{TPOT}  \\
\hline
Base + LoRA & 0 & 1x  \\
\hline
R + LoRA & 15 & 0.88x \\
\hline
R + LoRA & 20 & 0.81x \\
\hline
R + LoRA & 27 & 0.76x \\
\bottomrule
\end{tabular}
\caption{\small{TPOT variation of LLaMA-3-8B on WMT (left) and CNN/DM (right) for FiRST. The reported values are relative to the LoRA fine-tuned base model. Fine-tuning improves TPOT and quality significantly.}}
\label{tab:tpot_mini_8}
\end{table*}

\begin{table*}
\renewcommand{\arraystretch}{1.05}
\centering
\scriptsize
\begin{tabular}{ c |c |c|c}
\toprule
\textbf{Model Type} & \textbf{$\sim$ Skipping (\%)} & \textbf{Eng$\rightarrow$De}&\textbf{Eng$\rightarrow$Zh}\\
\textbf{} & \textbf{} & \textbf{TPOT}   &\textbf{TPOT}  \\
\hline
Base + LoRA & 0& 1x &1x \\
\hline
R + LoRA & 15& 0.90x &0.91x \\
\hline
R + LoRA & 25& 0.78x &0.75x \\
\hline
R + LoRA & 35& 0.69x &0.74x \\
\bottomrule
\end{tabular}
\hspace{1cm}
\begin{tabular}{ c  |c  |c l}
\toprule
\textbf{Model Type} & \textbf{$\sim$Skipping (\%)} & \textbf{CNN/DM}  &\\
\textbf{} & \textbf{} & \textbf{TPOT}  &\\
\hline
Base + LoRA & 0& 1x &\\
\hline
R + LoRA & 15& 0.88x &\\
\hline
R + LoRA & 24& 0.79x &\\
\hline
R + LoRA & 28& 0.77x &\\
\bottomrule
\end{tabular}
\caption{\small{TPOT variation of LLaMA-3.2-3B on WMT (left) and CNN/DM (right) for FiRST. The reported values are relative to the LoRA fine-tuned base model. Fine-tuning improves TPOT and quality significantly.}}
\label{tab:tpot_mini_3}
\end{table*}

\subsection{Detailed Result Table}\label{app:detailed_results}
Tables \ref{tab:translation_results_eng_de} and \ref{tab:translation_results_eng_zh} present the detailed results for the LLaMA-3.2-3B and LLaMA-3-8B models for an additional skipping percentage on Machine Translation Task. The results are reported using BLEU (BLEU-1, BLEU-2, BLEU) and COMET metrics, highlighting performance across different skipping percentages. Similarly, Table \ref{tab:cnn_dm_extended} presents cumulative results for both models reporting BERT F1, ROUGE-1 and ROUGE-L. Lastly, Table \ref{tab:extended_squad} presents Exact Match (EM) and F1 scores for both models for three skipping percentage variations.

\subsection{Generalizability and Reusability of Routers}
\label{app:generalizability and Reusability of Routers}
We also investigate whether a router trained on one dataset can be applied directly to another dataset for the same task, without any additional training. Analysis of layer‐skipping patterns suggests that models processing different datasets of the same task exhibit similar internal behavior, implying that a router learned on one dataset could be reused elsewhere. To validate this idea, we train routers on English→German translation data and test them on English→Chinese translation, and likewise train on SQuAD for question answering before evaluating on the Natural Questions dataset. The quantitative results of these cross-dataset evaluations are presented in Table~\ref{tab:nq_mt_side_by_side_cross}.
In the QA experiments, the router trained and tested on NQ itself (denoted as \textit{Original} in Table~\ref{tab:nq_mt_side_by_side_cross}) and the router trained on SQuAD but tested on NQ (\textit{Cross-dataset}) both skip a comparable fraction of layers. Despite being trained on different data, the SQuAD‐trained router achieves QA accuracy (EM score) very close to that of the NQ‐trained router, as detailed in Table~\ref{tab:nq_mt_side_by_side_cross}. This indicates that the essential decision patterns learned by the router transfer well across QA datasets when using a similar level of layer skipping.
A similar pattern emerges in machine translation. Whether trained on English→German or directly on English→Chinese data, routers that skip the same proportion of layers produce very similar translation quality metrics (BLEU and COMET in Table~\ref{tab:nq_mt_side_by_side_cross}) on the English→Chinese task. This holds across different model sizes and skip settings.
Together, these results demonstrate that routers learned on one dataset can be effectively reused on another dataset for the same task, as long as they are configured to skip a similar amount of computation. This reusability can lead to substantial savings in both training time and computing resources.

\begin{table*}
\centering
\tiny
% === LEFT: QA on NQ ===
\begin{tabular}{ c | c | c }
\toprule
\multicolumn{3}{c}{\textbf{Question Answering (Natural Questions)}} \\
\midrule
\textbf{Setting} & \textbf{Skip (\%)} & \textbf{EM} \\
\midrule
\multicolumn{3}{c}{\textbf{LLaMA-3.2-3B}} \\
\midrule
\textbf{Original}  & 10.32 & 22.55 \\
\textbf{Cross-Dataset}    & 12.46 & 20.87 \\
\textbf{Original}  & 18.03 & 17.51 \\
\textbf{Cross-Dataset}    & 19.04 & 16.93 \\
\midrule
\multicolumn{3}{c}{\textbf{LLaMA-3-8B}} \\
\midrule
\textbf{Original}  & 11.44 & 33.53 \\
\textbf{Cross-Dataset}    & 13.11 & 29.70 \\
\textbf{Original}  & 24.60 & 18.02 \\
\textbf{Cross-Dataset}    & 26.80 & 19.58 \\
\bottomrule
\end{tabular}
\hspace{1cm}
% === RIGHT: MT Eng–Zh ===
\begin{tabular}{ c | c | c | c }
\toprule
\multicolumn{4}{c}{\textbf{Machine Translation (Eng–Zh)}} \\
\midrule
\textbf{Setting} & \textbf{Skip (\%)} & \textbf{BLEU} & \textbf{COMET} \\
\midrule
\multicolumn{4}{c}{\textbf{LLaMA-3.2-3B}} \\
\midrule
\textbf{Original} & 14.69 & 0.37 & 0.55 \\
\textbf{Cross-Dataset}    & 16.57 & 0.37 & 0.45 \\
\textbf{Original} & 26.12 & 0.25 & 0.28 \\
\textbf{Cross-Dataset}    & 26.16 & 0.25 & 0.30 \\
\midrule
\multicolumn{4}{c}{\textbf{LLaMA-3-8B}} \\
\midrule
\textbf{Original}  & 16.46 & 0.25 & 0.43 \\
\textbf{Cross-Dataset}    & 14.70 & 0.37 & 0.49 \\
\textbf{Original}  & 27.73 & 0.25 & 0.25 \\
\textbf{Cross-Dataset}    & 30.00 & 0.25 & 0.24 \\
\bottomrule
\end{tabular}
\caption{%
Left: Exact‐match (EM) on Natural Questions (NQ) under two skip settings—\textit{Original} denotes routers trained and tested on the Natural Questions dataset, while \textit{Cross-dataset} denotes routers trained on SQuAD and tested on Natural Questions. Right: BLEU and COMET on English$\rightarrow$Chinese translation under the same variants, \textit{Original} denotes routers trained and tested on the English$\rightarrow$Chinese WMT direction, and \textit{Cross-dataset} denotes routers trained on English$\rightarrow$German and tested on English$\rightarrow$Chinese. This experiment evaluates the reusability and generalizability of routers across different datasets for the same task.%
}

\label{tab:nq_mt_side_by_side_cross}
\end{table*}

\end{document}